\documentclass[sigconf,screen]{acmart}

\usepackage{enumitem}
\setlist{leftmargin=5.5mm}

\usepackage{xcolor}
\definecolor{olivegreen}{rgb}{0, 0.6, 0}
\definecolor{redorange}{HTML}{FF5349}
\definecolor{blue(ncs)}{rgb}{0.0, 0.53, 0.74}
\definecolor{navy}{HTML}{273BE2}

\definecolor{black}{HTML}{000000}
\definecolor{white}{HTML}{ffffff}
\definecolor{color1}{HTML}{ACE5EE}
\definecolor{color2}{HTML}{0093AF}
\definecolor{color3}{HTML}{CC0000}
\definecolor{color4}{HTML}{0087BD}
\definecolor{color5}{HTML}{333399}
\definecolor{color6}{HTML}{20B2AA}

\usepackage{xspace}

\usepackage{verbatim}

\usepackage{multicol}
\usepackage{multirow}
\usepackage{makecell}
\usepackage{booktabs}

\usepackage{nicefrac}

\usepackage{algorithm}
\usepackage{algpseudocode}
\algnewcommand{\IfThen}[2]{
  \State \algorithmicif\ #1\ \algorithmicthen\ #2\ }
\algnewcommand{\IfThenElse}[3]{
  \State \algorithmicif\ #1\ \algorithmicthen\ #2\ \\ \algorithmicelse\ #3\ }

\usepackage[noabbrev, capitalize]{cleveref} 

\usepackage{marginnote}
\usepackage{tcolorbox}

\usepackage{pifont}
\newcommand{\cmark}{{\color{olivegreen}\ding{51}}}
\newcommand{\xmark}{{\color{red}\ding{55}}}

\newcommand{\thiswork}{GraNNDis\xspace}

\newcommand{\cobatch}{cooperative batching\xspace}
\newcommand{\CoBatch}{Cooperative Batching\xspace}
\newcommand{\Cobatch}{Cooperative batching\xspace}

\newcommand{\mempreload}{flexible preloading\xspace}
\newcommand{\Mempreload}{Flexible preloading\xspace}
\newcommand{\MemPreload}{Flexible Preloading\xspace}

\newcommand{\masking}{one-hop graph masking\xspace}
\newcommand{\Masking}{One-hop graph masking\xspace}
\newcommand{\MasKing}{One-Hop Graph Masking\xspace}

\newcommand{\intersampling}{expansion-aware sampling\xspace}
\newcommand{\Intersampling}{Expansion-aware sampling\xspace}
\newcommand{\InterSampling}{Expansion-aware Sampling\xspace}

\usepackage{tikz}
\newcommand*\circled[1]{\tikz[baseline=(char.base)]{
            \node[shape=circle,draw,inner sep=0.4pt, fill=black, text=white] (char) {#1};}}

\usepackage[pages=some]{background}

\AtBeginDocument{%
  \providecommand\BibTeX{{%
    \normalfont B\kern-0.5em{\scshape i\kern-0.25em b}\kern-0.8em\TeX}}}

\acmYear{2024}\copyrightyear{2024}
\setcopyright{rightsretained}
\acmConference[PACT '24]{The International Conference on Parallel Architectures and Compilation Techniques}{October 14--16, 2024}{Long Beach, CA, USA}
\acmBooktitle{The International Conference on Parallel Architectures and Compilation Techniques (PACT '24), October 14--16, 2024, Long Beach, CA, USA}
\acmDOI{10.1145/3656019.3676892}
\acmISBN{979-8-4007-0631-8/24/10}

\begin{document}

\title{\thiswork: Fast Distributed Graph Neural Network Training Framework for Multi-Server Clusters}
\author{Jaeyong Song}
\orcid{0000-0001-9976-7487}
\affiliation{%
  \institution{Seoul National University}
  \city{Seoul}
  \country{South Korea}
}
\email{jaeyong.song@snu.ac.kr}

\author{Hongsun Jang}
\orcid{0000-0003-4291-6124}
\affiliation{%
  \institution{Seoul National University}
  \city{Seoul}
  \country{South Korea}
}
\email{hongsun.jang@snu.ac.kr}

\author{Hunseong Lim}
\orcid{0009-0008-4945-1423}
\affiliation{%
  \institution{Seoul National University}
  \city{Seoul}
  \country{South Korea}
}
\email{hunseong.lim@snu.ac.kr}

\author{Jaewon Jung}
\orcid{0000-0003-0770-9277}
\affiliation{%
  \institution{Seoul National University}
  \city{Seoul}
  \country{South Korea}
}
\email{jungjaewon@snu.ac.kr}

\author{Youngsok Kim}
\orcid{0000-0002-1015-9969}
\affiliation{%
  \institution{Yonsei University}
  \city{Seoul}
  \country{South Korea}
}
\email{youngsok@yonsei.ac.kr}

\author{Jinho Lee}
\authornote{Corresponding author.}
\orcid{0000-0003-4010-6611}
\affiliation{%
  \institution{Seoul National University}
  \city{Seoul}
  \country{South Korea}
}
\email{leejinho@snu.ac.kr}


\begin{abstract}
\label{sec:abstract}
Graph neural networks (GNNs) are one of the rapidly growing fields within deep learning.
While many distributed GNN training frameworks have been proposed to increase the training throughput, they face three limitations when applied to multi-server clusters.
1) They suffer from an inter-server communication bottleneck because they do not consider the inter-/intra-server bandwidth gap, a representative characteristic of multi-server clusters.
2) Redundant memory usage and computation hinder the scalability of the distributed frameworks. 
3) Sampling methods, de facto standard in mini-batch training, incur unnecessary errors in multi-server clusters.

We found that these limitations can be addressed by exploiting the characteristics of multi-server clusters.
Here, we propose \emph{\thiswork}, a fast distributed GNN training framework for multi-server clusters.
Firstly, we present \emph{\MemPreload}, which preloads the essential vertex dependencies server-wise to reduce the low-bandwidth inter-server communications.
Secondly, we introduce \emph{\CoBatch}, which enables memory-efficient, less redundant mini-batch training by utilizing high-bandwidth intra-server communications.
Thirdly, we propose \emph{\InterSampling}, a cluster-aware sampling method, which samples the edges that affect the system speedup.
As sampling the intra-server dependencies does not contribute much to the speedup as they are communicated through fast intra-server links, it only targets a server boundary to be sampled.
Lastly, we introduce \emph{\MasKing}, a computation and communication structure to realize the above methods in multi-server environments.
We evaluated \thiswork on multi-server clusters, and it provided significant speedup over the state-of-the-art distributed GNN training frameworks.
\thiswork is open-sourced at \url{https://github.com/AIS-SNU/GraNNDis_Artifact} to facilitate its use.

\end{abstract}




\received{2024-04-01}
\received[accepted]{2024-06-27}


\begin{sloppypar}
\maketitle
\end{sloppypar}

\backgroundsetup{opacity=1, scale=1, angle=0, contents={
\begin{tikzpicture}[remember picture, overlay]
\node[anchor=north east, inner xsep=50pt, inner ysep=10pt] at (current page.north east) {
\href{https://www.acm.org/publications/policies/artifact-review-and-badging-current}{
\includegraphics[width=50pt]{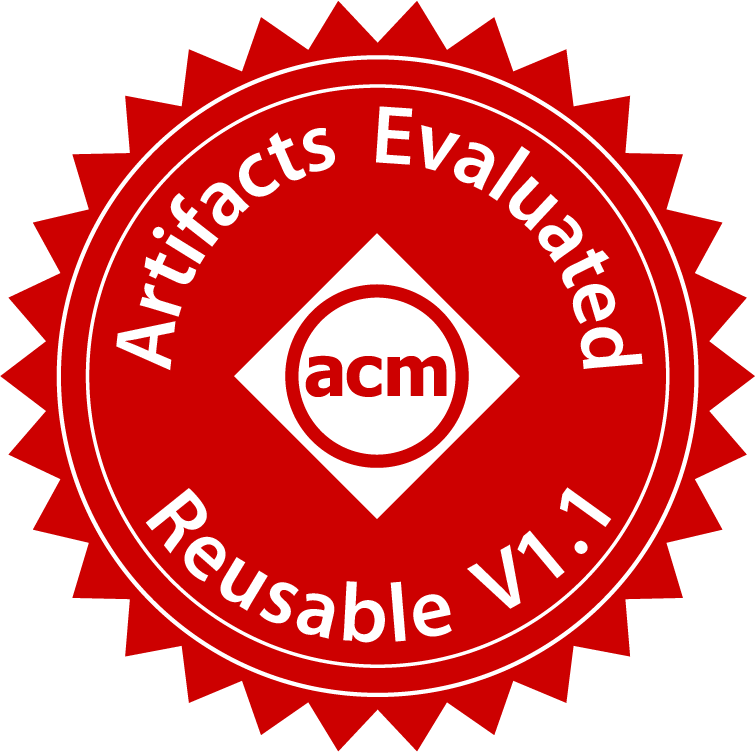}
\includegraphics[width=50pt]{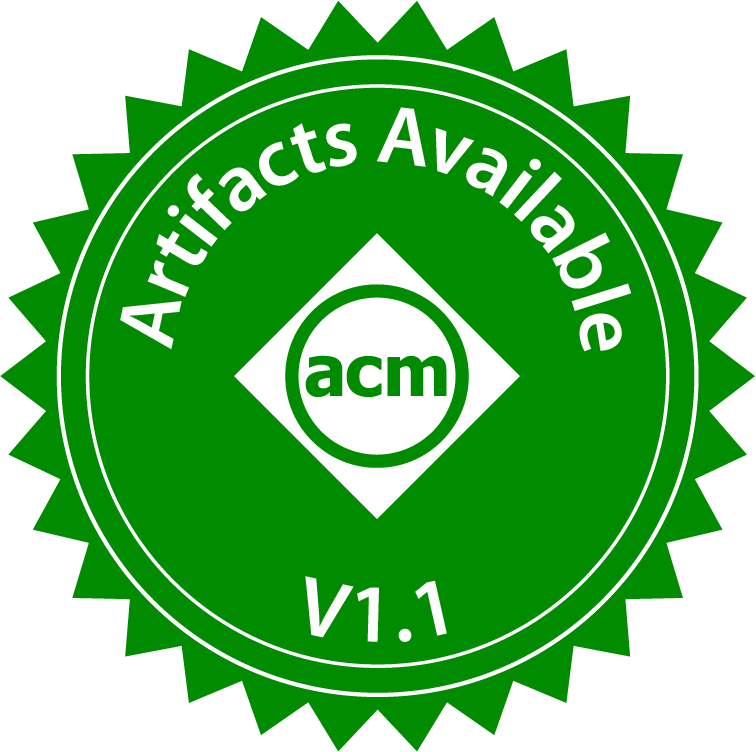}
\includegraphics[width=50pt]{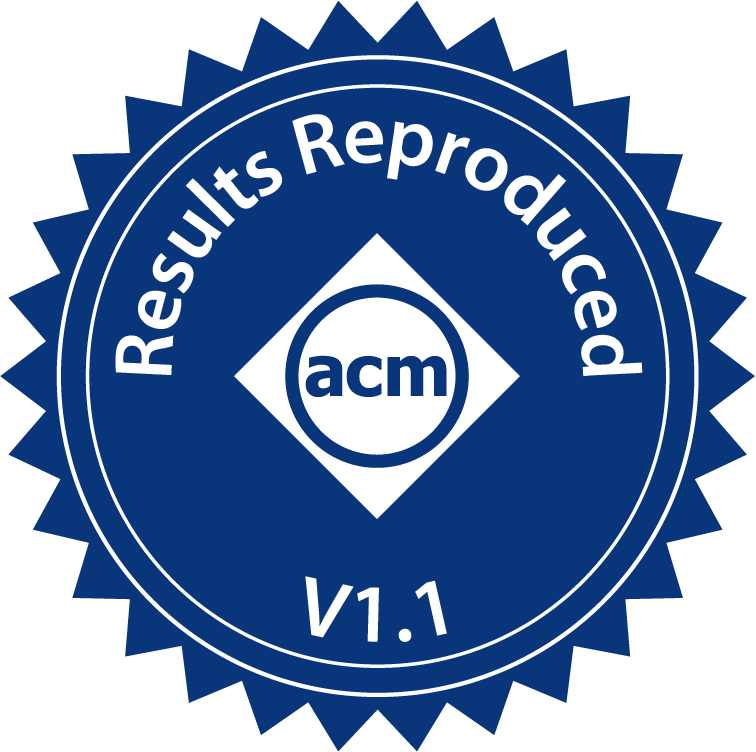}
}};
\end{tikzpicture}
}}
\BgThispage

\section{Introduction}

\label{sec:intro}

Over the last few years, graph neural networks (GNNs) have attracted increasing attention among various deep learning tasks.
As graphs can represent many types of non-structured data that do not fit into structured formats (e.g., images, or texts), they are widely used in various fields, such as social network user clustering~\cite{social_gnn}, protein analysis~\cite{protein_predict}, physics~\cite{physics_gnn}, and even traditional machine learning problems~\cite{image_gnn, nlp_gnn}.
Since the initial establishment of a few pioneering GNN algorithms~\cite{spectral, chebnet, gcn}, various algorithms have been proposed~\cite{graphsage, graphsaint, gat, gin} that capture different aspects of the knowledge within graph data.
The development of these algorithms was also accompanied by advancements in the training frameworks that provide higher throughput and convenient interfaces for implementation~\cite{pyg, dgl, roc}.

To achieve a higher training throughput, many distributed GNN training frameworks~\cite{roc, distdgl, distdglv2, agl, aligraph, neugraph, neutronstar, p3, pagraph, bgl} have been proposed.
Full-batch-based frameworks~\cite{roc, sancus, pipegcn, bns_gcn} try to reduce the communication bottleneck of distributed training.
Mini-batch-based frameworks~\cite{distdgl, distdglv2, salient, salient_pp} mitigate the redundant computation and memory usage of mini-batch training from the neighbor explosion through sampling approaches~\cite{graphsage, fastgcn, drop_edge, clustergcn, graphsaint, sampling}.

However, when conducting distributed GNN training on multi-server clusters, we found several challenges that must be addressed:
\begin{enumerate}
    \item \emph{Inter-Server Communication Bottleneck.} While some researches \cite{pipegcn, bns_gcn, sancus} address the communication bottleneck of distributed full-batch GNN training, they only focused on multi-GPU cases.
    As even high-speed inter-server links (around a few tens of GB/s)~\cite{infiniband} are two orders of magnitude lower than the GPU device memory bandwidth~\cite{hbm}, and several times slower than intra-server links~\cite{gpu_interconnect} in multi-server clusters, inter-server communication becomes the new bottleneck.
    \item \emph{Redundant Memory Usage/Computation.} Multi-server clusters are adopted for GNN training, especially when using large graphs~\cite{igb, ogb} or deep models~\cite{gcnii,deepgcn,deepergcn, revgcn}.
    With such cases, redundant memory usage and computation still exist even when sampling methods~\cite{graphsage, fastgcn, drop_edge, clustergcn, graphsaint, sampling, shadowsampling} are applied.
    \item \emph{Unnecessary Error from Sampling.} Existing sampling approaches \cite{graphsage,graphsaint,fastgcn,shadowsampling} are manipulating the whole graph structure.
    However, in multi-server clusters, sampling the intra-server edges (dependencies) does not contribute much to increasing throughput, as they are communicated through the fast intra-server links.
    Therefore, we need another sampling strategy to reduce the unnecessary information loss from a sampling.
\end{enumerate}

Fortunately, we identified that these challenges can be addressed by utilizing the opportunity provided by the characteristics of multi-server clusters.
Here, we propose \emph{\thiswork}\footnote{\thiswork comes from the Latin word \emph{grandis} meaning `large.' It is also a combination of Graph, NN, and Distributed.}, a fast distributed graph neural network training framework for multi-server clusters.

\begin{sloppypar}
Firstly, we propose \mempreload, in which each server preloads the vertex dependencies required and distributes them to the intra-server GPUs.
This strategy utilizes intra-server links instead of inter-server ones, thus reducing the inter-server communication bottleneck.
Since it requires additional memory, \mempreload dynamically selects the preloading hyperparameter considering the remaining server-wise aggregated GPU memory.
\end{sloppypar}

Secondly, we present \cobatch, which creates large-sized batches for mini-batch training by merging the GPU-wise mini-batches of existing frameworks.
While previous mini-batch methods suffer from a significant redundancy issue, \cobatch addresses it by aggregating multiple GPUs in a server with fast intra-server links.

Thirdly, we suggest \intersampling, a new sampling method suited for multi-server clusters.
Sampling methods mainly focus on manipulating the whole graph structure.
In multi-server clusters, we found that sampling the intra-server edges (dependencies) does not bring much speedup but could harm the convergence, as those edges are communicated through the high-speed intra-server links.
Therefore, \intersampling targets to sample only inter-server boundary and can mitigate the neighbor explosion with lower error.

\begin{sloppypar}
Lastly, we introduce \masking to realize a framework that considers multi-server clusters while supporting both full-batch and mini-batch training.
By server-wise subgraph extraction and dependency masking with one-hop graph-based data structure, \thiswork can support the above-proposed methods with high throughput.
Additionally, it aids \cobatch to orthogonally support existing sampling methods.
\end{sloppypar}

With the proposed strategies, we evaluated \thiswork on multi-server clusters.
\thiswork showed significant training throughput increases and comparable/lessened sampling errors compared to prior arts.

In summary, this paper makes the following contributions:
\begin{itemize}
    \item We identify the advantage of utilizing the characteristics of multi-server clusters to address the challenges in distributed GNN training.
    \item We propose \mempreload, which minimizes the inter-server communication bottleneck by utilizing the fast intra-server links.
    \item We present \cobatch, which significantly reduces the redundancy issue in mini-batch training by aggregating multiple GPUs in a server.
    \item We suggest \intersampling, a sampling strategy that considers the characteristics of multi-server clusters. 
    \item We introduce \masking; a computing and communication structure for training in multi-server clusters, to implement the above contributions.
    \item We evaluated GraNNDis extensively in multi-server clusters to show the speedup and usability compared to prior work.
\end{itemize}

\section{Background}
\label{sec:background}

\begin{figure}[t]
    \centering
    \includegraphics[width=.95\columnwidth]{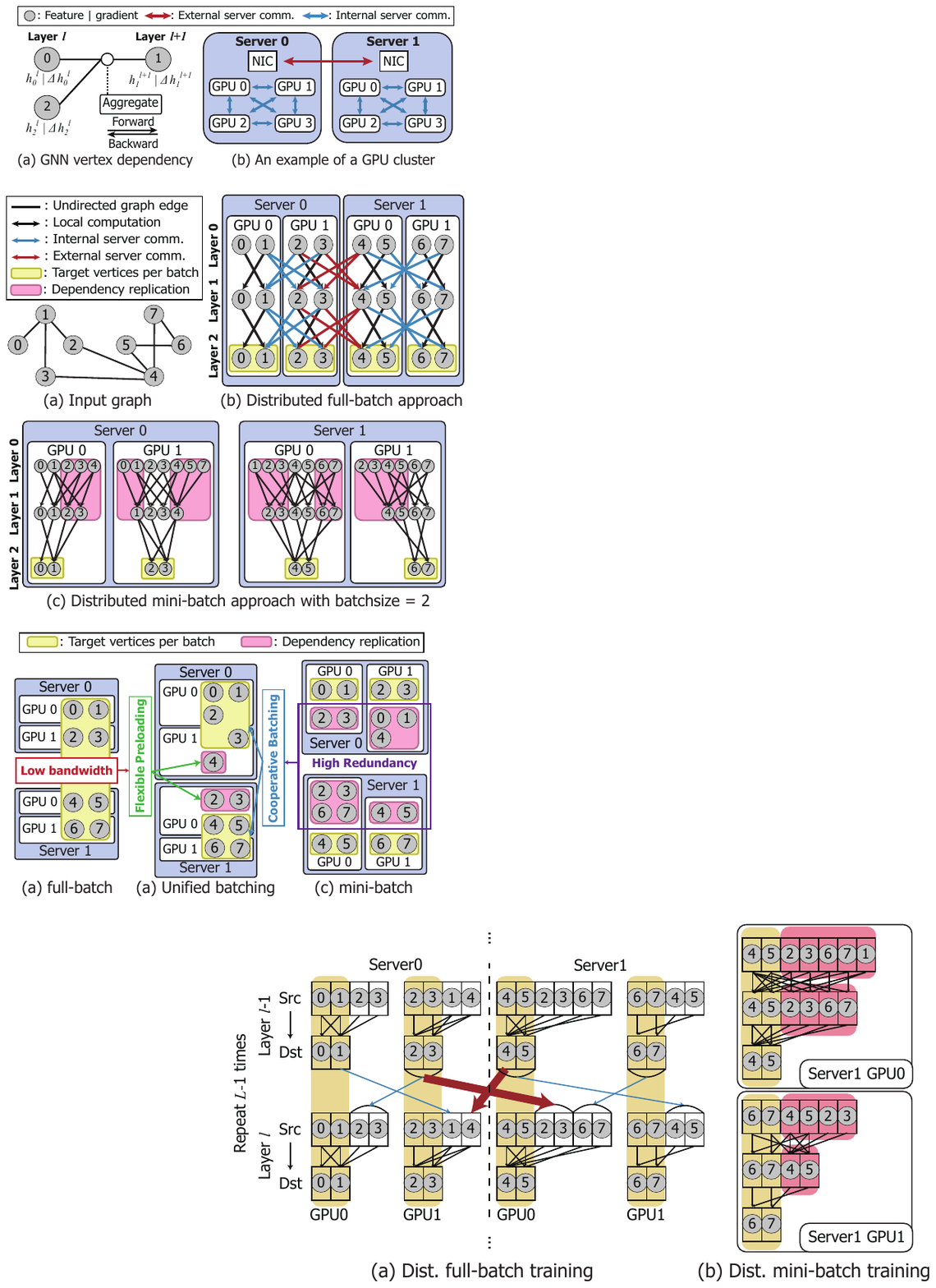}
    \vspace{-4mm}
    \caption{(a) An example of vertex dependency on neighbor vertices.
    (b) A multi-server cluster environment.
    }\vspace{-3mm}
    \Description[]{A background figure that shows an example vertex dependency and multi-server cluster environment.}
    \label{fig:background}
\end{figure}

\subsection{Graph Neural Network (GNN) Training}
\label{sec:gnn}
GNNs~\cite{gcn, gat, gin} show powerful potential for graph representation learning, and as such they are widely used for diverse tasks~\cite{detection_task1, deepgcn, deepergcn, revgcn}.
While transformer architectures~\cite{transformer, graphormer} have been widely used recently, GNNs are lighter and more specialized in learning graphs' topological characteristics than others.
Therefore, GNNs are preferred in real-time workloads~\cite{flowgnn, dgnn_booster}, data with irregular connectivity (e.g., social networks~\cite{social_gnn}) and molecular dynamics~\cite{protein_predict,gemnet,nequip}.
The advantages of GNNs can also be exemplified by a trend to utilize GNNs and transformers together~\cite{equiformer,equiformer_v2,longrange_graph_attention,grover}.
Modern GNN structures most commonly use neighbor dependency information ($\mathcal{N}$) to generate the final representation. 
Because of this,  
GNNs recursively require neighbor vertices' hidden vectors of the previous layer to generate the next representation of vertex $v$. 
This is called \emph{vertex dependency} in GNN training, as depicted in \cref{fig:background}a.
Every layer ($l$) of GNN  generates the next hidden vector $h_v^{l}$ for each target vertex $v$ 
by aggregating hidden vectors of $v$'s neighbors $\mathcal{N}(v)$,
multiplying weights of layer ($W^{l}$), and lastly applying activation function ($\sigma$).
The aggregated vector ($h^{l}_{\mathcal{N}(v)}$) is often combined with the previous hidden vectors 
of target vertices $h^{l-1}_{v}$.
This can be formulated as follows:
\begin{equation}
\resizebox{\columnwidth}{!}{%
$\begin{aligned}
h^{l}_{\mathcal{N}(v)}=\mathit{AGGREGATE}(\{h^{l-1}_{u}| u \in \mathcal{N}(v)\}),
h^{l}_{v}=\overrightarrow{F}^{l}(h^{l-1}_{v}, h^{l}_{\mathcal{N}(v)}),\notag
\label{eq:background}
\end{aligned}$
}
\end{equation}
where $AGGREGATE(\cdot)$ is an aggregation function, usually the summation of all the neighbor hidden vectors. $\overrightarrow{F}^{l}(\cdot)$ comprises the multiplication of weights ($W^{l}$) followed by the activation function ($\sigma$).
This processing of dependency on neighbor vertices makes GNNs unique compared to other neural network models~\cite{lenet, seq2seq}. 
Most GNNs mainly suffer from this aggregation process of intermediate representation, thus causing a severe slowdown in training~\cite{gnnear}. 

Traditional GNNs~\cite{gcn, graphsage} often use three to five layers, but deeper GNNs~\cite{deepgcn, deepergcn, gcnii} with residual connections show improved accuracy in various areas.
Deep GNNs cause over-smooting~\cite{yang2020revisitingoversmoothing, gcnoversmooth}, but many works~\cite{deepgcn, deepergcn, gcnii} minimize it through normalization/residual connections, achieving higher accuracy~\cite{protein_predict, ogb}.
In such deep GNNs, the training becomes much harder because they often use tens or hundreds of layers, requiring much higher memory requirements and much heavier computations.
A representative problem is a \emph{neighbor explosion}, which is a result of the increased \textit{Message Flow Graph} (MFG) size by the number of layers. 
It is becoming more common to use multiple accelerators (e.g., GPUs) to address such challenges.

\subsection{Distributed GNN Training}
\label{sec:dist_training}

Two representative distributed training methods are used to mitigate the outbursting memory usage of GNN training: full-batch and mini-batch training.
Full-batch training is memory-hungry but reaches high final accuracy, while mini-batch training requires less memory and converges faster, but the final accuracy could be lower. Thus, both approaches are widely used.

\begin{figure}
    \centering
    \includegraphics[width=.95\linewidth]{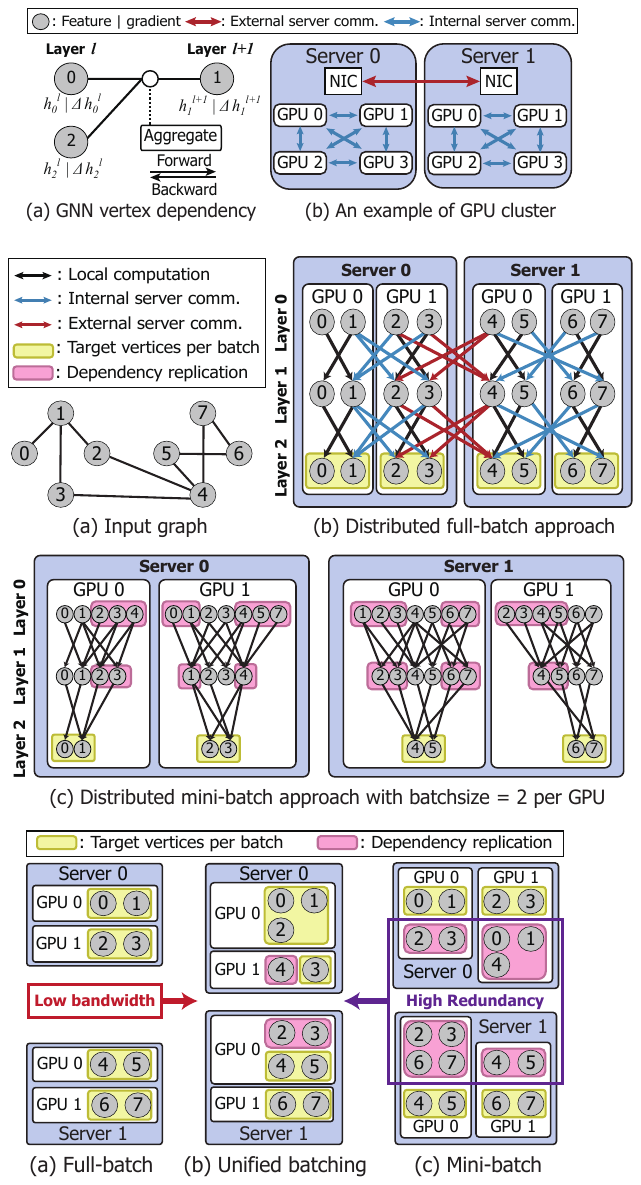}
    \vspace{-3mm}
    \caption{Forward-pass of a 2-layer GNN with two servers, each with two GPUs. (a) An example of an input graph. (b) An example of a full-batch approach. (c) An example of a mini-batch approach with batch size=2 per GPU. 
    }\vspace{-4mm}
    \Description[]{Input graph example and existing approaches. Overall figures of full-batch and mini-batch training.}
    \label{fig:dist_training}
\end{figure}

\emph{Full-batch training}~\cite{roc, pipegcn, bns_gcn, neutronstar} uses the whole graph representation for every parameter update.
As depicted in~\cref{fig:dist_training}b, all vertex dependencies must be fully computed.
Therefore, an iteration (model parameter update) essentially becomes an epoch.
For this, all the hidden vectors for every layer have to be loaded on the GPUs' memory, so multiple GPUs are typically required for large graphs~\cite{leskovec, ogb} and deep GNNs~\cite{deepgcn, deepergcn} in a distributed manner.

In distributed full-batch training, inter-GPU communications are essential for retrieving intermediate vertex representations from other GPUs, as illustrated by the \textbf{\color{blue}blue} and \textbf{\color{red}red} arrows in~\cref{fig:dist_training}b.
The overhead from this communication is significant, so it generally uses a partitioning algorithm~\cite{metis} to minimize the communication volume when partitioning a graph for GPUs.
However, most existing works~\cite{neugraph, agl, roc} do not consider multi-server cluster settings and heavily rely on slow external communications.
This work provides higher throughput than previous works by considering a characteristic of multi-server clusters that a difference exists between the inter- and intra-communication bandwidth, such as~\cref{fig:background}b.

\emph{Mini-batch training}~\cite{dgl, distdgl, distdglv2, clustergcn, graphsaint} is a technique used when the GPU memory capacity is insufficient for full-batch training. 
It uses a batch concept to mitigate the memory capacity issue.
It packs the partial dependencies into MFGs for the fixed number of target vertices to mitigate the heavy memory overheads, as illustrated with MFGs in~\cref{fig:dist_training}c.
By adjusting batch size, training can be performed with flexible memory requirements,
at the cost of redundant memory usage and computation among mini-batches.

Mini-batch training can be easily extended to distributed training~\cite{distdgl, distdglv2, p3}, as depicted in~\cref{fig:dist_training}c. 
Each GPU loads all the vertex dependencies required to compute the feature of target vertices, which means the distributed mini-batch approach does not communicate for fulfilling vertex dependencies between GPUs. 
After each GPU processes its mini-batch, the weight update is performed collectively by sharing the gradients among the GPUs.
In this case, the total batch size is \textit{the mini-batch size} $\times$ \textit{the number of GPUs}.
Compared to full-batch training, an epoch requires several iterations to handle all vertices.
However, it scales poorly because the GPU-wise mini-batch leads to extreme redundancy in memory usage and computations, as shown in \cref{fig:dist_training}c (shaded pink), and demonstrated in~\cref{sec:motiv}.

\subsection{Sampling Methods}
Because of the high memory requirements and heavy processing overhead of dependencies in GNNs, many sampling methods~\cite{graphsage, fastgcn, drop_edge, clustergcn, graphsaint, sampling, shadowsampling} are actively being explored.
They can be roughly divided into \emph{layer-wise sampling} and \emph{subgraph sampling} approaches.
Layer-wise sampling approaches~\cite{graphsage, fastgcn, drop_edge} apply sampling algorithm layer by layer, typically constraining the maximum degree of edges called \textit{fanout}.
Subgraph sampling approaches~\cite{clustergcn, graphsaint, shadowsampling} sample the subgraph from the original graph and execute forward and backward propagation on the sampled subgraph only.
Mini-batch training approaches~\cite{dgl, distdgl, distdglv2, p3} are commonly merged with sampling methods to minimize their redundancy.
These sampling methods manipulate the whole graph structure and could be unsuitable for graphs with a high average degree.
\thiswork introduces a new sampling method advantageous for high average degree graphs considering the characteristics of a multi-server cluster environment.

\section{Motivational Study}
\label{sec:motiv}

In the previous sections, we discussed existing distributed GNN training methodologies and their limitations.
\cref{tab:baselines} directly summarizes the prior arts compared to \thiswork.
\thiswork is an internal/external server communication-aware and redundancy-aware framework.
Additionally, through \masking, \thiswork supports both full-batch and mini-batch training.

\begin{figure}[t]
\centering
 \includegraphics[width=\columnwidth]{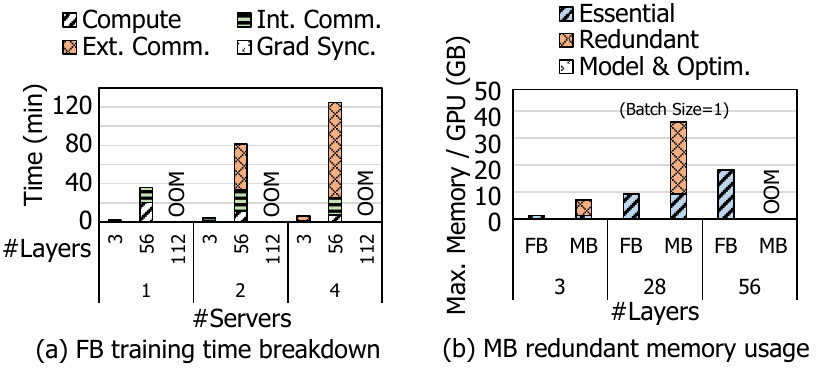}
 \vspace{-7mm}
 \caption{Motivational experiments.}
 \vspace{-4mm}
 \Description[]{Motivational experiments.}
 \label{fig:motiv}
\end{figure}

In the present section, we show a motivational study to emphasize the key insights of \thiswork.
\cref{fig:motiv} shows the result of a motivational study, where the limitations of current distributed full-batch~\cite{pipegcn} (FB) and mini-batch~\cite{salient_pp} (MB) GNN training can be observed.
We trained shallow (3-layer) from deep (28/56/112-layer) GNNs with residual connections~\cite{deepgcn} on various datasets.
All GNN training is conducted on a four-server cluster, where each server has four RTX A6000 GPUs.
For the detailed setup, please refer to~\cref{sec:env}.

\textbf{\textit{Insight \#1:} Some works considered multi-server environments but with naive expansion without considering the intra-/inter-server bandwidth gap.}
\cref{fig:motiv}a shows the training time breakdown of the state-of-the-art full-batch training framework~\cite{pipegcn} when increasing the number of servers with the Reddit dataset.
We break down the training time into four: computation, inter-server communication, intra-server communication, and parameter gradient synchronization.
Ideally, using multiple servers should speed up the training, but due to the inter-server communication bottleneck, distributed training is unscalable and even slows down.
This is because inter-server communication is much slower than intra-one, and the currently used full-batch GNN training frameworks do not consider this difference.

\textbf{\textit{Insight \#2:} Mini-batch-based works suffer from multi-server unsuitable structure (i.e., MFG).}
As illustrated in \cref{fig:motiv}a, training deep 112-layer setting faces out-of-memory (OOM).
A representative strategy for addressing this issue is mini-batch training.
However, mini-batch training suffers from significant redundancy even in the state-of-the-art distributed mini-batch training framework~\cite {salient_pp}, which invalidates the strategy.
\cref{fig:motiv}b illustrates such redundancies in various models with different numbers of layers on the Reddit dataset.
We applied GraphSAGE~\cite{graphsage} sampling with 25 fixed fanouts.
Even though the sampling~\cite{graphsage} is adopted, massive redundancy still exists due to the neighbor explosion.
We started from a batch size of 1K per GPU in a 3-layer case, but with 56 layers, the batch size of only one vertex consumes more memory than the full-batch case, and no batch size fits with 56 layers, even with a minibatch size of 1.
There are multiple reasons, but we find the primary reason to be the redundancy caused by the mini-batches being formed in individual GPUs. 

\begin{table}[t]
    \centering
    \caption{Comparison of \thiswork to Prior Art}
    \label{tab:baselines}
    \vspace{-3mm}
    \resizebox{\columnwidth}{!}{%
    \setlength{\tabcolsep}{3pt}
    \begin{tabular}{ccccccc}
    \toprule
         \multirowcell{3}{Method}&\textbf{\makecell{\multirowcell{3}{Multi-Server\\Cluster\\Aware}}}&\multicolumn{2}{c}{Full-Batch (FB)} & \multicolumn{2}{c}{Mini-Batch (MB)}  & \textbf{\makecell{\multirowcell{3}{Support \\ (Both?)}}}\\
         \cmidrule(lr){3-4}
         \cmidrule(lr){5-6}
         & & \makecell{Comm.\\Aware} & \textbf{\makecell{Int./Ext.\\Aware}} & \makecell{Multi-\\Server} & \textbf{\makecell{Batch Limit \\ (Redundancy Aware)}} &  \\
        \midrule

         CAGNET~\cite{cagnet} & \xmark & \xmark & \xmark & - & - & FB (\xmark) \\
         Sancus~\cite{sancus} & \xmark & \cmark & \xmark & - & - & FB (\xmark)\\
         PipeGCN~\cite{pipegcn} & \xmark & \cmark & \xmark & - & - & FB (\xmark)\\
         BNS-GCN~\cite{bns_gcn} & \xmark & \cmark & \xmark & - & - & FB (\xmark)\\
         \midrule
          PaGraph~\cite{pagraph} & \xmark & - & - & \xmark & a GPU (\xmark) & MB (\xmark) \\
        DSP~\cite{dsp} & \xmark & - & - & \xmark & a GPU (\xmark) & MB (\xmark)\\
        P$^{3}$~\cite{p3} & \xmark & - & - & \cmark & a GPU (\xmark) & MB (\xmark)\\
        MariusGNN~\cite{marius} & \xmark & - & - & \xmark & a GPU (\xmark) & MB (\xmark)\\
        WholeGraph~\cite{wholegraph} & \xmark & - & - & \cmark & a GPU (\xmark) & MB (\xmark)\\
        Quiver~\cite{quiver} & \xmark & - & - & \cmark & a GPU (\xmark) & MB (\xmark)\\
        SALIENT++~\cite{salient_pp} & \xmark & - & - & \cmark & a GPU (\xmark) & MB (\xmark)\\
         \midrule
         \textbf{\makecell{\thiswork\\(Proposed)}} & \cmark & \cmark & \cmark & \cmark & \makecell{Multiple GPUs (\cmark)} & \makecell{FB+MB (\cmark)}\\
        \bottomrule
    \end{tabular}
    } \vspace{-3mm}
\end{table}

\begin{sloppypar}
\textbf{\textit{Insight \#3:} Existing samplings (layer-wise/subgraph) mostly target single GPU environments, which incurs unnecessary information loss.}
In \cref{fig:motiv}a, we can figure out that the communication time of inter-server edges (dependencies) consumes a significantly larger portion than the intra-server ones.
This implies that sparsifying (sampling) the intra-server edges does not contribute much to a speedup, however, existing samplings do not consider this characteristic from multi-server clusters.    
\end{sloppypar}

\section{Notations and Performance Model}
For analysis, we set some notations to explain the latency model of distributed GNN training.
$N_s$ and $N_g$ indicate the total number of servers and the number of GPUs per server, respectively.
$V$ means the total number of vertices in a given graph and
$E$ is the number of edges.
$C$ is the computational throughput measured in vertices per second.
$B_{\mathit{intra}}$ and $B_{\mathit{inter}}$ are intra-/inter-server bandwidths split per GPU, which are also measured in vertices per second.
$L$ is the total number of GNN layers, which means that the model extracts the information from $L$-hop neighbors.
In the analysis, we assume that hidden dimension sizes and vertex feature sizes are equal in every layer for convenience.

We formulate the training time of the previous full-graph training method ($T_{\mathit{prev}}$) as follows, concentrating on the communication of the vertex dependencies that cannot be overlapped. 
{
\begin{align}
T_{\mathit{prev}}&=\frac{V}{N_sN_gC}  
        + \frac{E}{N_sN_g} \biggr[\frac{(\nicefrac{(N_g-1)}{N_gN_s})}{B_{\mathit{intra}}}
        + \frac{(1-1/N_s)}{B_{\mathit{inter}}}\biggr]. \label{eq:prev}
\end{align}
}
The first term of~\cref{eq:prev} presents the total compute time of the previous full-graph training.
The second term calculates the total communication time, which is the summation of intra- and inter-server communication time.
One observation from the communication term is that inter-server communication has larger numerators (more inter-server GPUs) and smaller denominators (lower inter-server bandwidth).
This makes inter-server communication become the bottleneck under the latency model.
The parameter gradient synchronization by all-reduce communication is omitted because it is almost entirely overlapped by computation through gradient bucketing~\cite{pytorch}.

\section{\thiswork Framework}
\label{sec:granndis}

Using the characteristics of multi-server clusters, we introduce three schemes of \thiswork to address the issues from the motivation study.
We propose methods to gain speedup under memory budget for full-batch training (\cref{sec:preload}) and to generally support diverse graph sampling methods~\cite{graphsage,graphsaint} on large batches for mini-batch training (\cref{sec:cobatch}).
In addition, we suggest another novel sampling method for multi-server clusters, which gives great efficiency in speed and achieved accuracy (\cref{sec:sampling}).

\begin{figure}[t]
\centering
\includegraphics[width=0.88\columnwidth]{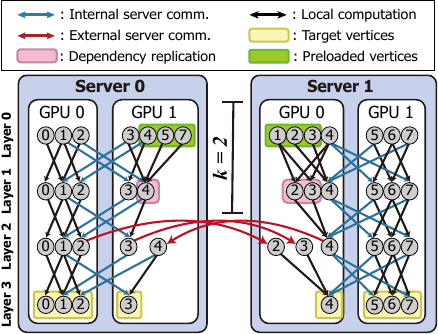}
\vspace{-3mm}
 \caption{\Mempreload mitigates the inter-server communication bottleneck using fast intra-server links. The green area represents the preloaded initial graph features.}
 \vspace{-5mm}
 \Description[]{Overall figure of \mempreload.}
 \label{fig:preload}
\end{figure}

\subsection{\MemPreload}
\label{sec:preload}

\begin{tcolorbox}[top=0mm, bottom=0mm]
\begin{itemize}[leftmargin=*]
\item \textbf{Key Idea}: Minimize inter-server communication within the memory budget in full-batch distributed training.
\end{itemize}
\end{tcolorbox} 

In~\cref{sec:motiv}, we showed that previous distributed GNN training methods suffer from low scalability from not considering the inter-/intra-server bandwidth difference of multi-server clusters.
We propose \emph{\mempreload} for full-batch training, which considers this difference to reduce inter-server communications.
As discussed earlier, the inter-server connections (e.g., Ethernet, Infiniband) are typically an order of magnitude slower than the intra ones (e.g., PCIe, NVLink) in multi-server clusters.
Therefore, it is important to avoid using inter-server connections as much as possible.

To minimize inter-server communication, the vertex dependencies should be handled in another way.
\Mempreload uses a preloading strategy to fulfill such dependencies.
We can choose to include the remote vertices in the extracted graph (replication) instead of having to communicate, similar to mini-batching methods~\cite{distdgl, distdglv2}.
However, the power-law distributions of real-world graphs~\cite{leskovec} would easily cause the extracted graph to be very large, which diminishes the point of distributed training from a memory capacity perspective.

\Mempreload attempts to avoid such problems but still minimizes inter-server communications, as depicted in \cref{fig:preload}.
Starting from layer 0, the same one-hop graph structure is used for a set of consecutive $k$ layers applying \masking, which will be discussed in \cref{sec:masking}. 
On the next layer (i.e., layer $L-k+1$), the one-hop graph is reshaped, which is then reused in the succeeding layer.
This reduces inter-server communication by replicating vertices at a minimal amount of reshaping cost.
We choose the largest $k$ to make the GNN fit on the available memory.

\cref{fig:preload} illustrates an example where $k=2$ is used for a 3-layer GNN.
The vertices are split into two servers such that vertices 4-7 belong to server 1 (the inner vertices of server 1).
Note that we use imbalanced partition among GPUs for a clear view, but balanced partitioning~\cite{metis} is used in practice, because the neighbors expand in all directions in real-world graphs.
one-hop graph masking is used but omitted from the illustration.
In layer 0, one-hop neighbors are preloaded (shaded green), reducing the inter-server communication.
For layers from one to $k-1$, \masking is needed to adapt to consider the exact dependency.
After layer $L-k+1=2$, a new (smaller) one-hop graph is used, which allows inter-server communications (vertex 2, 3, and 4) in the subsequent layers.
However, most communications are intra-server, not severely slowing down the system.

\Mempreload incurs some redundant computation across servers as a cost for reduced inter-server communication.
However, it is advantageous because the devices' computational throughput far exceeds that of the inter-server communication in recent multi-server clusters.
We approximately formulate such a trade-off with a performance model.
From~\cref{eq:prev}, we can derive the training time and the expected speedup using \mempreload.
Here, we assume that the dependency graph size grows at the rate of $D^\alpha$ every layer, where $D$ is the average degree of vertices, and $0<\alpha<1$ is a graph-dependent per-layer expansion factor to represent overlapping neighbors.
Also, we set $k=L$ for brevity.
The inter-server bandwidth ($B_{inter}$) is replaced by the computation of preloaded vertices. 
In addition, the preloaded vertices increase the amount of intra-server communications.
With these considerations, the training time changes as follows: 
\begin{align}
T_{\mathit{preload}}=\frac{VD^\alpha L}{N_sN_gC},
        + \frac{ED^\alpha L}{N_sN_g} \biggr[\frac{\nicefrac{(N_g-1)}{N_g}}{B_{\mathit{inta}}}\biggr] , \label{eq:preload}
\end{align}
\begin{equation}
1 < {D^\alpha L} \leq N_sN_g.
\end{equation} 
The condition for $T_{\mathit{preload}}$ to be shorter than $T_{\mathit{prev}}$ from \cref{eq:prev} is  $T_{\mathit{preload}}-T_{\mathit{prev}} < 0$.
Making some rough approximations as $\nicefrac{1}{B_{\mathit{internal}}} \approx 0$ and $\nicefrac{1}{N}\approx0$, the condition becomes
\begin{equation}
\frac{VD^\alpha L}{N_sN_gC} - \frac{V}{N_sN_gC}  
        - \frac{E}{N_sN_g} \frac{1}{B_{\mathit{inter}}}<0. 
        \end{equation}
Because $ E=V\cdot D$  by definition,
        \begin{equation}
        \frac{C}{B_{\mathit{inter}}} > \frac{(D^\alpha L-1)}{D} . \label{eq:compare}
        \end{equation}
\cref{eq:compare} suggests how much gap is needed between computational throughput and communication bandwidth for \mempreload to gain speedup. 
Using $D^\alpha=2$ and $D=10$ for a three-layer GNN ($L=3$), \mempreload has an advantage when the processing rate (vertex / second) of computation is at least half that of the inter-server communication, which is the usual case in cluster environments.

\begin{figure}[t]
\centering
\includegraphics[width=0.95\columnwidth]{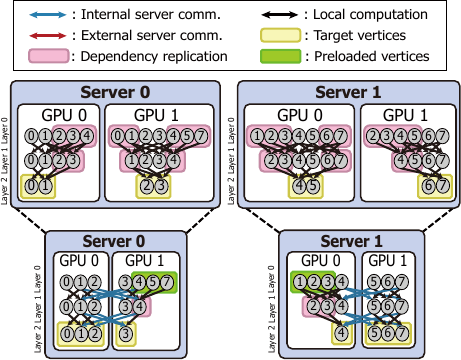}
\vspace{-2mm}
 \caption{In the conventional mini-batch approach, the memory size of a single GPU bounds the mini-batch size. \Cobatch enables larger batches while reducing redundancy using fast intra-server links.}
 \vspace{-5mm}
 \Description[]{Concpetual figure of \cobatch.}
 \label{fig:cobatch}
\end{figure}

\subsection{\CoBatch}
\label{sec:cobatch}

\begin{tcolorbox}[top=0mm, bottom=0mm]
\begin{itemize}[leftmargin=*]

\item \textbf{Key Idea}:
Reduce redundancy and support sampling techniques in mini-batch distributed training.
\end{itemize}
\end{tcolorbox}

In mini-batch training framework, the vertex dependencies are fetched to individual GPUs. 
This strategy results in a lot of redundant memory usage and greatly limits the size of a mini-batch, as depicted in \cref{fig:background}c.
We propose \cobatch to mitigate such limitation of the previous strategy with the aid of \masking, which will be discussed in \cref{sec:masking}.
With \cobatch, to suppress redundancy and enable a much larger batch size, we cooperate the GPUs within a server as if they were a single large GPU, as illustrated in~\cref{fig:cobatch}.
We choose the target vertices to form a \textit{co-batch} and form an MFG of them.
While existing mini-batch training frameworks fetch MFGs to each GPU, \cobatch fetches co-batches to each server.
Then, each co-batch is distributed to the internal GPUs of each server.

The application of \cobatch has a clear advantage of substantially less redundant memory and, thus, a larger batch size.
It also significantly reduces the redundant computation of mini-batch training.
By enlarging a batch size limit, we could address the outbursting memory usage of recent GNNs due to an increase in the number of layers and graph size growth.
The disadvantages are increased intra-server communication and the possibility of an imbalanced workload between workers. 
However, neither is a severe overhead because the intra-server bandwidth is very high, and the redundancy is significantly reduced.

One remaining issue is supporting sampling techniques. 
As sampling approaches~\cite{graphsage,graphsaint} are almost de facto standard to mini-batch training, \cobatch also supports existing sampling methods by \masking.
For layer-wise samplings~\cite{graphsage, fastgcn, drop_edge}, the maskings are generated per layer to conduct layer-wise aggregation of them.
While existing subgraph sampling methods~\cite{graphsaint, clustergcn} construct subgraphs in a GPU-wise manner, they are identical to the unmasked version of \masking when extending them to a server-wise manner, which will be described in \cref{sec:masking}.

\subsection{\InterSampling}
\label{sec:sampling}
\begin{tcolorbox}[top=0mm, bottom=0mm]
\begin{itemize}[leftmargin=*]
\item \textbf{Key Idea}:
Apply sampling only in the expansion region that affects the system speed.
\end{itemize}
\end{tcolorbox}

\begin{figure}[t]
\centering
\includegraphics[width=.85\columnwidth]{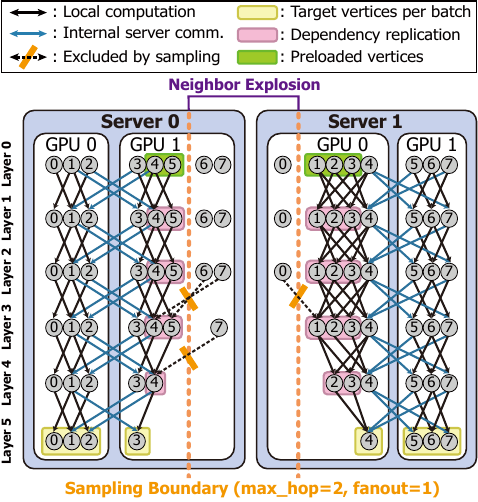}
\vspace{-3mm}
 \caption{
 \Intersampling applies a sampling on the server boundary to mitigate the neighbor explosion.
}\vspace{-5mm}
\Description[]{Overall figure of \intersampling.}
 \label{fig:sampling}
\end{figure}

As discussed in \cref{sec:background}, sampling methods~\cite{graphsage, sampling, graphsaint, clustergcn, drop_edge} are popular techniques in both full-batch and mini-batch training, which aim to prevent the neighbor explosion~\cite{leskovec, clustergcn, bns_gcn}.
Meanwhile, the implication behind the huge speedup of \mempreload and \cobatch is that computation and intra-server communication are at least several times faster than inter-server communication.
Considering these, we observe that the sparsification of dependencies inside a GPU, or among GPUs in the same server does not contribute much to the speedup while they lose the same amount of information in a graph.
In such regard, prior sampling methods could be unsuitable for graphs with high average degrees (e.g., Reddit, average degree 492) because they sample the whole graph structure (e.g., with fanout 25). 

Therefore, we propose \intersampling, which aligns with the lessons from the performance implications in multi-server clusters.
Instead of sampling the whole graph structure, \intersampling only targets to sample the server boundary vertices in a multi-server cluster.
Only from the server boundary vertices, we define maximum hops ($h$) and maximum fanouts ($f$) towards the external vertices in remote servers for sampling, and maintain all other dependencies (i.e., inner vertices) untouched.
As we will show, it adds a great amount of speedup compared to other methods, at a comparable or better accuracy.

\subsubsection{Sampling Details}
\cref{fig:sampling} illustrates the detailed sampling methodology of \intersampling.
For simplicity and visibility, we assume that vertices 0-3 and vertices 4-7 are dedicated to servers 0 and 1, respectively.
When the vertices for each server that are not in inner vertices are preloaded for full-/mini-batch training, \intersampling applies constraints to decide whether a certain vertex is included.

We provide an example of $h=2$ and $f=1$.
For \intersampling, we select external vertices that are at maximum a two-hop distance ($h=2$) from the server's inner vertices.
Moreover, \intersampling restricts the hop traversing to find only one external neighbor ($f=1$).
When applying the fanout, to consider the importance of each vertex, \intersampling selects the vertices that have many connections (degrees) to the unselected vertices. 
Therefore, the vertices from 4 to 7 are to be preloaded by server 0 without any sampling, but \intersampling only decides to preload the vertices 4 and 5.
When the traversal is finished, a subgraph is generated from the traversed vertices, and the internal dependencies of the inner vertices remain unsampled as a whole.
 We empirically used the settings of $h=1$ and $f=15$, which aligns with prior works~\cite{gnnautoscale, lmc} that emphasize the importance of 1-hop dependency.
 We observe that the number of inner vertices (at least 10K) is sufficiently large, and the min-cut algorithm (i.e., METIS~\cite{metis}) for dedicating inner vertices helps the sampling to contain the whole graph's structural information.
We provide the sensitivity of the tunable parameters in~\cref{sec:sensi}.
The cost of \intersampling is formulated as below:
\begin{align}
T_{\mathit{sampling}}&=\frac{Vm^\alpha k}{N_sN_gC} 
        + \frac{Em^\alpha k}{N_sN_g} \biggr[\frac{(N_sN_g - 1)}{B_{\mathit{intra}}}\biggr], \label{eq:sampling}
\end{align}
\begin{equation}
 1 < m^{\alpha} k \ll D^{\alpha} L \leq N_sN_g.
\end{equation} 
While this preserves the number of edges in the perspective of the whole graph, it has the effect of reducing the growth to $m^{\alpha k}$, much smaller than $D^{\alpha L}$ since $m\ll D$ and $k\ll L$.

\subsubsection{Approximate Error Bound Analysis}
\label{sec:err_bound}

\begin{sloppypar}
We approximately compare \intersampling with the layer-wise sampling~\cite{graphsage} and the subgraph sampling~\cite{clustergcn} as follows.
By comparing the approximate error bounds, we found that \intersampling provides a smaller error than the previous subgraph sampling method~\cite{clustergcn} and is advantageous for high average-degree graphs than the layer-wise sampling~\cite{graphsage}.
\end{sloppypar}

\textbf{Error bound of samplings.}
We apply Theorem 2 of~\cite{gnnautoscale} for the average pooling and replace historical error terms ($\epsilon$) with sampling error terms ($\epsilon_{samp}$) as the historical and sampling~\cite{graphsage, graphsaint} errors both affect the node embeddings in each layer.
$emb_{v}^{L}$ is a full-batch embedding of $v$ after passing $L$ layers.
The closeness of the approximated embedding to the full-batch embedding can be represented by $||\tilde{emb}_{v}^{L} - emb_{v}^{L}|| \geq 0$.
The error of the sampled embedding from the approximated embedding for each layer $l$ is $0 \leq ||\tilde{emb}_{v, sampled}^{(l)} - \tilde{emb}_{v}^{(l)}|| \leq \epsilon_{samp}^{(l)}$.
Let $AGGREGATE(\cdot)$ and $\overrightarrow{F}^{l}(\cdot)$ be Lipschitz continuous with constants $c_1$ and $c_2$, and set $c_1c_2$ as $C$.
According to~\cite{gnnautoscale}, the final error of a vertex $v$ is approximately bounded by
\begin{align}
    ||\tilde{emb}_{v}^{L} - emb_{v}^{L}|| \lesssim \Sigma_{l=0}^{L-1}{\epsilon_{samp}^{(l)}C^{L-l}}. \label{eq:error_lemma_layers}
\end{align}
The error from a layer ($l-1$) propagates to the next layer ($l$) with the constant $C$.
Therefore, $C^{L-l}$ intuitively shows that the errors from earlier layers propagate and accumulate to deeper layers.

\textbf{Layer-wise sampling~\cite{graphsage}.} We can set the error of the layer-wise fanout sampling as $\epsilon_{fanout}$ and replace $\epsilon_{samp}$ of \cref{eq:error_lemma_layers}.
$|V|$ indicates the total number of vertices.
When we set the average error among vertices from a fanout sampling as $\bar{\epsilon}_{fanout}$, the total maximum error bound of all vertices using a fanout sampling method can be calculated as follows using \cref{eq:error_lemma_layers}:
\begin{align}
     ERR_{fanout}^{max} = \Sigma_{l=0}^{L-1}{\bar{\epsilon}_{fanout}^{(l)}{|V|}{C^{L-l}}}.\label{eq:approx}
\end{align}

\textbf{\Intersampling (Ours).} By changing \cref{eq:approx} for \intersampling, we can additionally figure out the importance of not sparsifying the dependencies among GPUs in the same server.
\thiswork generates a huge batch at a server level.
In this case, a relatively small portion of inner vertices requires outer-boundary vertices at each hop.
To consider such an effect of a large batch, we can introduce coefficients $\alpha$ and $\beta$ ($\alpha+\beta=1$).
$\alpha$ is the ratio of the total number of inner vertices not requiring outer-boundary vertices to the number of preloaded vertices ($\alpha > 0$): therefore, $\alpha$ becomes large as batch size increases.
We set the error from an inner vertex as $\epsilon_{in}$, which is close to $0$ because inner vertices do not use sampled information for each hop.
Meanwhile, $\beta$ is the ratio of the number of boundary vertices to the number of replicated vertices ($\beta > 0$).
We set the error from a boundary vertex as $\epsilon_{out}$, which is a small value with the aid of a min-cut algorithm such as METIS~\cite{metis}.
When setting the coefficients and errors to derive the maximum error bound of \intersampling in \cref{eq:approx},
\begin{equation}
\resizebox{\columnwidth}{!}{%
$\begin{aligned}
    ERR_{EAS}^{max} = \Sigma_{l=0}^{L-1} {(\alpha^{(l)}\bar{\epsilon}_{in}^{(l)}+\beta^{(l)}\bar{\epsilon}_{out}^{(l)})}{|V|}{C^{L-l}} \simeq \Sigma_{l=0}^{L-1} {\beta^{(l)}\bar{\epsilon}_{out}^{(l)}}{|V|}{C^{L-l}}.\label{eq:err_eas}
\end{aligned}$
}
\end{equation}
$\beta$ is roughly inversely proportional to the batch size (or partition size), and therefore the error is small for larger batch sizes.
However, most existing sampling, such as fanout sampling (\cref{eq:approx}), has no such effects because it naively restricts the fanout of \textit{all vertices}.
We compare the actual error of \intersampling ($ERR_{EAS}^{actual}$) with the fanout sampling ($ERR_{fanout}^{actual}$) and analyze the empirical values of $\beta$ in \cref{sec:acc_compare}.

\textbf{Subgraph sampling~\cite{clustergcn}.} We can apply \cref{eq:err_eas} to subgraph sampling methods~\cite{clustergcn, graphsaint, shadowsampling} that use a smaller GPU-wise subgraph mini-batch, because \intersampling can be seen as a generalization of subgraph sampling.
In those cases, the mini-batch size becomes extremely small (a hundred to a few thousand), so $\beta$ is close to $1$ ($\beta \simeq 1$).
Also, it drops more outer boundary information, so $\bar{\epsilon}_{subg}$ would be larger than $\bar{\epsilon}_{out}$.
Therefore, the above makes the maximum error always larger than that of \intersampling as follows:
\begin{align}
    ERR_{subg}^{max} \simeq \Sigma_{l=0}^{L-1} {(1\cdot\bar{\epsilon}_{subg}^{(l)})}{|V|}{C^{L-l}} > ERR_{EAS}^{max}. \label{eq:err_subg}
\end{align}
We also compare the actual errors of the subgraph sampling method ($ERR_{subg}^{actual}$) to other methods in \cref{sec:acc_compare}.

\section{\MasKing}
\label{sec:masking}

\begin{figure*}[t]
\centering
 \includegraphics[width=0.95\textwidth]{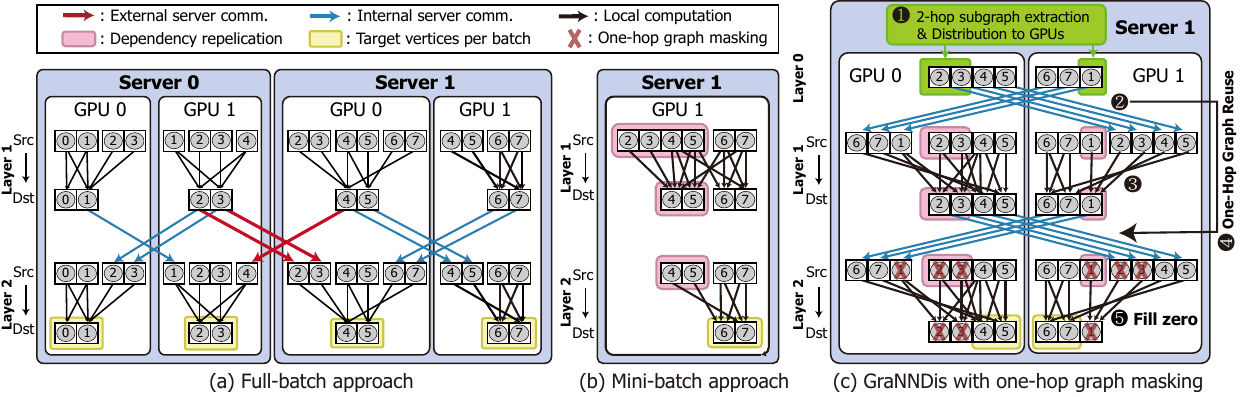}
 \vspace{-3mm}
 \caption{Baseline approaches and overall structure of \thiswork with \masking on a multi-server cluster.}
 \vspace{-3mm}
 \Description[]{Overall figure of \masking.}
 \label{fig:structure}
\end{figure*}

\begin{tcolorbox}[top=0mm, bottom=0mm]
\begin{itemize}[leftmargin=*]
\item \textbf{Key Idea}: Support both full-/mini-batch training to efficiently use intra-server links of multi-server clusters.
\end{itemize}
\end{tcolorbox}

The methods of \thiswork utilize the fast intra-server bandwidth of recent multi-server clusters while avoiding the slower inter-server communications on both full-batch and mini-batch training.
As conceptual methods, we propose \mempreload and \cobatch.
However, realizing such a unified framework with existing work is difficult, because of the underlying difference in the software structure.
Therefore, we introduce \masking, an efficient data structure, and software architecture to realize it.

\cref{fig:structure}a depicts the commonly used structure of a state-of-the-art full-batch training~\cite{pipegcn, bns_gcn}.
They are based on a one-hop graph structure (src$\xrightarrow[]{}$dst) to minimize random accesses and avoid frequent graph reshaping. 
Between each layer, the vertex features are fetched from remote GPUs, either by inter- (red) or intra- (blue) server communications.
To avoid heavy random access overhead, a separate contiguous memory region is assigned for every peer GPU.
Then, the required vertex features are transferred among GPUs in an all-to-all~\cite{mpi} pattern, where the vertices are locally sorted.
This way, the accesses are piecewise sequential in both the communication and computation.
Furthermore, the same one-hop graph structure is used in every layer, avoiding repetitive reshaping operations.

On the other hand, mini-batch training frameworks~\cite{distdgl, distdglv2, salient, salient_pp} take a completely different approach as shown in \cref{fig:structure}b.
Because all dependent neighbors are included in the MFG, there is no structure for communication.
Furthermore, each layer has a differently shaped graph, which prohibits naive adoption of one-hop graph structures from \cref{fig:structure}a.
While it is doable in principle, switching to different subsets of vertices with distinct ordering would cause severe latency overhead.

To address aforementioned issues, \thiswork uses a novel strategy of \emph{\masking}.
The key idea is to reuse the same one-hop graph structure, but apply different masking every layer to support varying vertex subsets and dependency handling.
This approach takes the benefit from both sides of full-batch and mini-batch training.
The partially sequential accesses and graph reuse property of the full-batch training are preserved, while the effective structure can freely vary with high flexibility.

\cref{fig:structure}c shows the procedure and structure of \thiswork with \masking based on the example graph. 
\circled{1} As the first step, it extracts selected neighbors from multi-hop dependency starting from the inner vertices of a server.
Then, the extracted vertices are distributed to the GPUs in a server.
For example, if we choose to extract entire 2-hop neighbors from the inner vertices 4-7, the vertices 1-7 are extracted and partitioned to GPUs (vertices 2-5 to GPU 0 and the rest to GPU 1)
\circled{2} For each layer, similar to the distributed full-batch training (\cref{fig:structure}a), each GPU receives the required vertex features.
In this example, because we chose 2-hop neighbors for the 2-layer GNN, only the intra-server communication remains, and all of the slow inter-server dependencies are already replicated in the extracted graph.
\circled{3} Then, the next vertex features are computed from the received features.
\circled{4} In the next layer, the same one-hop graph and communication pattern are reused.
After then, with \masking, we generate a mask for the unneeded vertices,  which fills zeros to the features.
\circled{5} As a result, we can generate the final outputs (logits) of the inner vertices considering the exact dependency.

\section{Implementation Details}
\label{sec:impl}

We implement \thiswork based on vanilla~\cite{pipegcn}
which significantly improves the throughput of~\cite{roc} and achieves state-of-the-art performance.
It uses GLOO~\cite{gloo}, which does not support GPU RDMA features; therefore, we reimplemented it with NCCL~\cite{nccl}.
Further, while the original method~\cite{pipegcn} uses custom parameter gradient synchronization, we modified it to use the PyTorch DistributedDataParallel (DDP) package to provide overlapping gradient synchronization.
These optimizations show significant speedup, so we set this optimized version~\cite{pipegcn} as our full-batch training baseline.

For parallelization, \thiswork uses subgraphs generated by \mempreload and \cobatch according to the batching scheme (full-/mini-batch).
The subgraphs are distributed to each server, where \masking is utilized for intra-server parallelization among GPUs.
\Masking further creates subgraphs for preloading and partitions the server-dedicated subgraphs.
For this purpose, we make a `head group' comprising the first process of each server, and this head group handles those procedures.
Additionally, \thiswork does not employ pipelining because pipelining for GNN training often indicates asynchronous staleness with lower accuracy and is seldom utilized in practice.

\section{Evaluation}
\label{sec:eval}

\subsection{Experimental Environment}
\label{sec:env}

\textbf{Cluster environment and model configuration.}  ~\cref{tab:environment} lists our cluster environments.
We use clusters equipped with four RTX A6000 GPUs, an AMD EPYC 7302 CPU, and a 512GB system memory.
Our cluster has Infiniband QDR (32 Gbps) for external connections between servers, and NVLink is used for internal connections.
We trained GraphSAGE~\cite{graphsage} for shallow GNNs as a default.
For deep GNNs, we used ResSAGE+~\cite{deepergcn} with GraphSAGE as a graph convolution.
We also trained GCN~\cite{gcn} for some evaluations.
Note that for full-batch training, GraphSAGE and ResSAGE+ do not use the fanout sampling of GraphSAGE but only adopt the model structure.
We used 64 as the default hidden dimension size in all experiments.

\begin{table}[t]
\footnotesize
\centering
 \def\arraystretch{0.95}
 \caption{Experimental Environment}\label{tab:environment}
 \vspace{-3mm}
{
\begin{tabular}{cccc}
\toprule
\multirowcell{6}[-0.4ex]{\textbf {HW}} 
& \multirowcell{4}{Server} 
& GPU & 4$\times$ NVIDIA RTX A6000 \\
&& CPU & 1$\times$ EPYC 7302, 16C 32T \\
&& Memory & 512GB DDR4 ECC \\
&& Int. connect & NVLink Bridge~\cite{nvbridge} \\
\cmidrule(lr){3-4}
& \multirowcell{2}{Cluster}
& \#Servers & 4 \\
&& Ext. connect & Infiniband QDR (32Gbps) \\

\midrule

\multirowcell{8}[-0.8ex]{\textbf {SW}} 
& \multirow{4}{*}{Common} 
 & Python & 3.10\\
& & PyTorch  & 1.13 \\
& & CUDA & 11.6.2 \\
& & NCCL & 2.10.3 \\

\cmidrule(lr){3-4} 
& \multirow{5}{*}{GNN}
& DGL & 0.9.1~\cite{dgl} \\
&& Model & \makecell{GraphSAGE~\cite{graphsage}, \\ ResSAGE+~\cite{deepergcn}, GCN~\cite{gcn}}\\
&& Hidden dim. & 64 \\
&& Task & Node classification \\

 \bottomrule
\end{tabular}
} \vspace{-3mm}
\end{table} 

\begin{table}[t]
    \centering
    \caption{Graph Datasets}
    \label{tab:datasets}
    \vspace{-3mm}
    \resizebox{\columnwidth}{!}
    {
    \setlength{\tabcolsep}{3pt}
    \begin{tabular}{ccccccccc}
    \toprule
         \multirowcell{2}{Size}& \multirowcell{2}{Name} &\multicolumn{3}{c}{Dataset Info.} & \multicolumn{3}{c}{Hyper-parameter} \\
         \cmidrule(lr){3-5}
         \cmidrule(lr){6-8}
         & & \makecell{\#Nodes} &  \makecell{\#Edges} & \makecell{Feat. size} &\makecell{lr} & \makecell{Dropout} & \makecell{\#Epochs} \\
         \midrule
         \multirowcell{3}{Small$\sim$\\medium} & ogbn-arxiv~\cite{ogb} & 0.17M & 1.2M & 128 & 0.01 & 0.5 & 1000 \\
          & ogbn-products~\cite{ogb} & 2.45M & 61.9M & 100 & 0.003 & 0.3 & 1000 \\
         & Reddit~\cite{reddit} & 0.23M & 114.6M & 602 & 0.01 & 0.5 & 1000 \\
         \cmidrule(lr){0-7}
         \multirowcell{3}{Large$\sim$\\Hyperscale} & IGB (S)~\cite{igb} & 1M & 12.1M & 1024 & 0.01 & 0.5 & 1000 \\
         & IGB (M)~\cite{igb} & 10M & 120.1M & 1024 & 0.01 & 0.5 & 1000 \\
         & ogbn-papers100M~\cite{ogb} & 111M & 1.6B & 128 & 0.01 & 0.5 & 1000 \\
        \bottomrule
    \end{tabular}
    } \vspace{-3mm}
\end{table}

\begin{figure*}[t]
\centering
 \includegraphics[width=.96\textwidth]{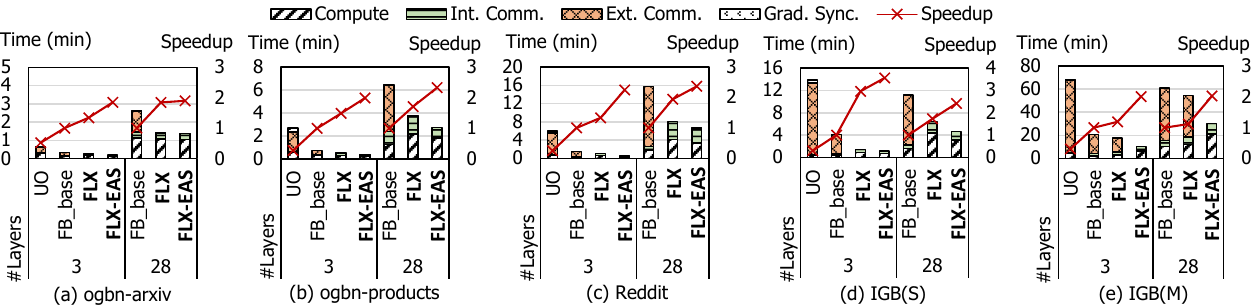}
 \vspace{-3mm}
 \caption{Full-batch training time breakdown and speedup of the unoptimized/optimized baselines (UO, FB\_base), \mempreload (FLX), and \intersampling (FLX-EAS) in shallow to deep GNNs.}
 \vspace{-3mm}
 \Description[]{Full-batch speedup comparison.}
 \label{fig:fb_speedup}
\end{figure*}

\textbf{Datasets.}
We chose six datasets; ogbn-arxiv (Arxiv, AX), ogbn-products (Products, PD), Reddit~\cite{reddit} (RD), igb-small (IGB (S)), igb-medium (IGB (M))~\cite{igb}, and ogbn-papers100M (Papers)~\cite{ogb}.
The characteristics of datasets are summarized in~\cref{tab:datasets}.
Arxiv is a citation network where each edge indicates a source paper citing a destination paper.
Products represents a co-purchasing network, and each node represents a product in Amazon.
Reddit consists of nodes representing posts in the online forum of Reddit.
IGB and Papers are also citation networks whose sizes are orders of magnitude larger than other datasets.
We mainly used the hyperparameters chosen in~\cite{pipegcn, igb}.
As Papers is a representative hyper-scale dataset, we dedicated a separate section in~\cref{sec:hypers}.

\textbf{Baselines.} We mainly chose two representative state-of-the-art baselines in our experiments, PipeGCN~\cite{pipegcn} and SALIENT++~\cite{salient_pp}.
PipeGCN~\cite{pipegcn} is a state-of-the-art full-batch training framework.
We used the vanilla version of~\cite{pipegcn} (i.e., without the pipeline that involves staleness), which is essentially a refactored version of existing full-batch training.
We optimized the vanilla~\cite{pipegcn} with NCCL, showing a much faster training time than the original version.
We set this as a full-batch training baseline and denoted it as FB\_base.
For the pipelined~\cite{pipegcn}, we tested it together with other full-batch-based staleness~\cite{sancus} and sampling~\cite{bns_gcn} methods in~\cref{sec:stale_sampling}.
SALIENT++~\cite{salient_pp} is a state-of-the-art distributed mini-batch training framework, and it mandates neighborhood sampling in each training iteration.
It significantly improves the training throughput of DistDGL~\cite{distdgl, distdglv2} with sampling/computation overlapping and aggressive caching.
We used SALIENT++ as a mini-batch training baseline and denoted it as MB\_base.
For further validation, we also reported the throughput of DistDGL~\cite{distdgl}.
However, DistDGL is significantly slower than the others, so we reported it only in~\cref{fig:mb_speedup}.
We trained all datasets for 1000 epochs with full-batch training baseline~\cite{pipegcn} and \thiswork.
With mini-batch training baselines, we set the total epoch of 30 and the batch size per GPU as 1024 and weak-scaled it, following the common practices from~\cite{salient_pp, distdgl, distdglv2}.
We tried to enlarge a batch size of mini-batch training baselines for a fair comparison with \thiswork.
However, it only worsens the sampler overhead of baselines and brings a slowdown to them.
Therefore, we settled to follow the optimized practices of their original papers~\cite{salient_pp, distdgl, distdglv2}.
Note that we applied layer-wise sampling~\cite{graphsage} for mini-batch baselines following the common practice.

\subsection{Training Speedup and Breakdown}
\label{sec:speedup}

\textbf{Full-batch training.} \cref{fig:fb_speedup} shows the speedup of \thiswork in full-batch training of GNNs compared to the unoptimized (UO) and optimized full-batch training baseline~\cite{pipegcn} with our optimizations (FB\_base).
We chose 3 and 28 layers, which are widely used as the number of layers for shallow~\cite{gcn, graphsage} and deep GNNs~\cite{deepgcn, deepergcn}.
For \intersampling (EAS), we used 1-hop and 15-fanout as default settings if not otherwise stated.
We verified the speedup from the optimization discussed in \cref{sec:impl}.
Our optimization of using NCCL (UO$\xrightarrow{}$FB\_base) provides 1.92-4.00$\times$ speedup, as illustrated in \cref{fig:fb_speedup}.
The unoptimized baseline fails to run deep GNNs due to its misuse of the system resources (i.e., sockets), so we reported its shallow layer case.

In all GNNs from shallow to deep layers, \mempreload (FLX) provides a speedup over the optimized baseline FB\_base, as shown in~\cref{fig:fb_speedup}.
This speedup comes from minimizing inter-server communication by preloading the vertex dependencies and sharing them within a server.
In FB\_base, inter-server communication consumes 44.07-85.81\% of the total training time.
\Mempreload (FLX) reduces this huge portion and provides 1.11-2.96$\times$ speedup.
In IGB (M), there is less chance to preload all vertex dependencies due to the memory limit, so \mempreload provides a relatively smaller, but still significant speedup of 1.19$\times$ and 1.11$\times$ in 3- and 28-layer GNNs, respectively.

\begin{figure*}[t]
\centering
 \includegraphics[width=\textwidth]{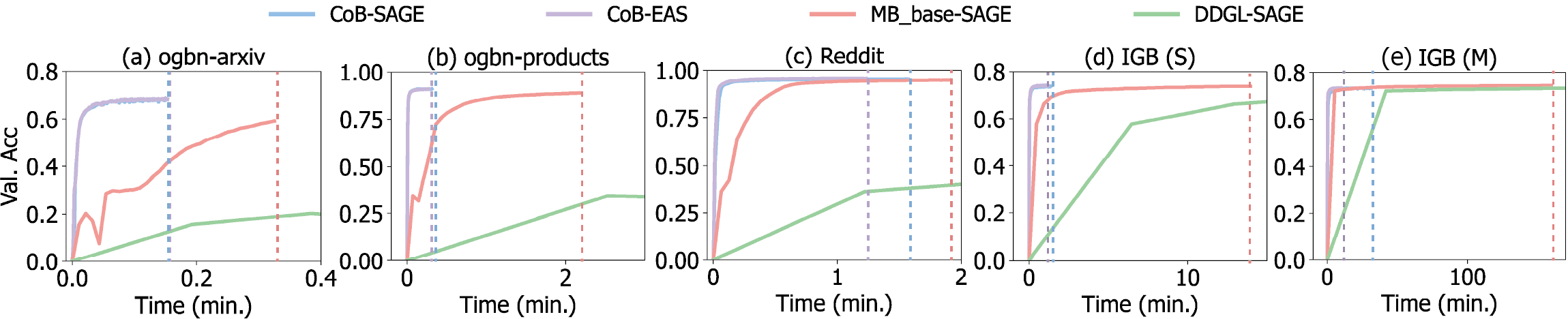}
 \vspace{-7mm}
 \caption{Mini-batch training time-to-accuracy curves of DistDGL (DDGL), SALIENT++ (MB\_base), and \cobatch (CoB) on various datasets. The dotted vertical lines depict the end of training.}
 \vspace{-3mm}
 \Description[]{Mini-batch speedup comparison.}
 \label{fig:mb_speedup}
\end{figure*}

\Intersampling (FLX-EAS) further accelerates the training by reducing the redundant computation and internal communication from the neighbor explosion.
For example, in the Reddit dataset with a 28-layer case, sole \mempreload already provides 1.94$\times$ speedup, but \intersampling addresses the neighbor explosion and shows 2.37$\times$ speedup.
For other datasets, when \intersampling is adopted, it provides a significant 1.85-3.56$\times$ speedup in shallow to deep GNNs.
It is worth noting that applying \intersampling yields comparable or better accuracy. We further discuss it in \cref{tab:acc_compare} and \cref{sec:acc_compare}.

\textbf{Mini-batch training.} \cref{fig:mb_speedup} shows time-to-accuracy plots of the proposed \cobatch (CoB) and SALIENT++~\cite{salient_pp}  (MB\_base), a famous state-of-the-art distributed mini-batch training method.
We trained the default shallow model, 3-layer 64-hidden size GraphSAGE.
We additionally examined DistDGL~\cite{distdgl} (DDGL), another popular distributed mini-batch training framework.
As mini-batch training methods generally mandate sampling methods, we adopted the fanout sampling from 25-fanout GraphSAGE~\cite{graphsage} to MB\_base and DDGL.
For a fair comparison, we also applied 25-fanout GraphSAGE to the proposed CoB (CoB-SAGE).
Additionally, we reported the time-to-accuracy results when \intersampling was adopted by CoB (CoB-EAS).

CoB-SAGE provides superior speedup over the baseline MB\_base of 1.62-12.69$\times$.
DDGL suffers from severe sampler overhead and is much slower than MB\_base and \cobatch.
For better visibility, we omitted the full training curve of DDGL.
MB\_base tends to show much significant redundancy in larger datasets with higher average degrees, so CoB-SAGE provides 12.69$\times$ speedup in the IGB~(M) dataset.
When comparing CoB-EAS with CoB-SAGE, CoB-EAS generally showed faster training time, showing up to 1.27$\times$ speedup in the Reddit dataset.
We omit the results from deep layers because MB\_base and DDGL yield out-of-memory errors either from the GPU or from the sampler in most datasets.
As we will show in the next, \thiswork allows the handling of deep layers.

\subsection{Hyper-Scale Dataset and Deeper Models}
\label{sec:hypers}

\begin{figure}[t]
\centering
 \includegraphics[width=\columnwidth]{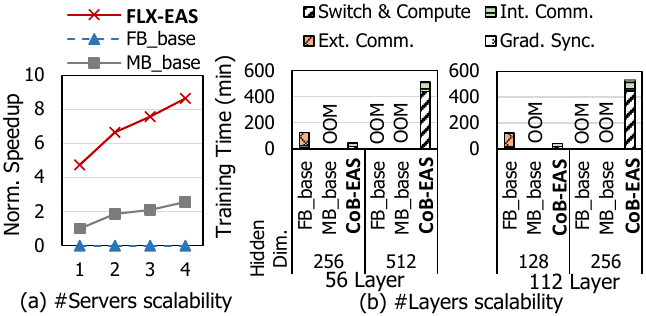}
 \vspace{-7mm}
 \caption{Scalability on the number of servers and layers.}
 \vspace{-3mm}
 \Description[]{Scalability test with a hyper-scale dataset.}
 \label{fig:hypers}
\end{figure}

To observe the scalability of \thiswork on a hyper-scale dataset, we trained ogbn-papers100M (Papers).
The results are shown in~\cref{fig:hypers}a. 
We used 3-layer 64-hidden size GraphSAGE and normalized all results to the training time of SALIENT++ running on a single server.
FB\_base fails to train from OOM issues, even with 16 GPUs from 4 servers.
SALIENT++ (MB\_base) is able to train the model with a smaller batchsize and aggressive sampling, but it does not scale well due to large redundancy.
On the other hand, \thiswork at full-batch mode using \intersampling (FLX-EAS) is able to train the model with good scalability in the number of servers.
FLX-EAS provides a huge speedup ranging from 3.37$\times$ to 4.74$\times$ compared to the baseline with the same number of servers.
Further, it achieves 60.84\% accuracy, while MB\_base got 60.58\%.
This supports that \intersampling provides stable convergence.

In \cref{fig:hypers}b, we demonstrate the benefit of \thiswork by providing \cobatch and \intersampling on deeper GNNs of 56- and 112-layers with Reddit dataset, and the hidden size ranging from 128 to 512.
In our cluster environment, both the FB\_base and the MB\_base baselines often fail from out-of-memory errors.
This is a natural phenomenon in FB\_base because it requires enough GPU memory to store all vertex features from all layers.
In such cases, MB\_base is expected to be an alternative, which allows training with smaller mini-batches.
Unfortunately, MB\_base failed to train them even at the mini-batch size of 1, despite our faithful effort to save memory space such as checkpointing~\cite{pytorch}.
On the other hand, \thiswork generates a batch at a server level with aggregated GPUs, which can mitigate heavy memory overhead of deeper and wider GNNs with less redundancy.
As a result, \thiswork in mini-batch mode with \intersampling (CoB-EAS) is the only choice to train the Reddit dataset in a 112-layer and 256-hidden dimension size with the cluster under test.
Moreover, when the model fits the memory with FB\_base, \thiswork provides speedups of 2.92$\times$ and 2.91$\times$ in 56- and 112-layer GNNs, respectively.

\subsection{Mini-Batch Sampling Methods on \thiswork}
As \thiswork supports most existing sampling methods with \cobatch, we compare their training time and accuracy implemented on \cobatch.
\cref{fig:orth_hidden}a shows the training time breakdown and accuracy of \cobatch when applying subgraph sampling (GraphSAINT~\cite{graphsaint}, SAINT), layer-wise sampling (GraphSAGE~\cite{graphsage}, SAGE), and \intersampling (EAS).
As SAINT removes any neighbors outside the target vertices, it provides a shorter training time but shows some accuracy degradation.
On the other hand, SAGE and EAS both consider further hops and fanouts of target vertices and show longer training time while achieving higher accuracy.
With \cobatch, applying GraphSAGE (SAGE) shows significantly less training time than MB\_base (in \cref{fig:mb_speedup}) from less redundancy.

\subsection{Sensitivity Studies}
\label{sec:sensi}

 \begin{figure}[t]
\centering
 \includegraphics[width=\columnwidth]{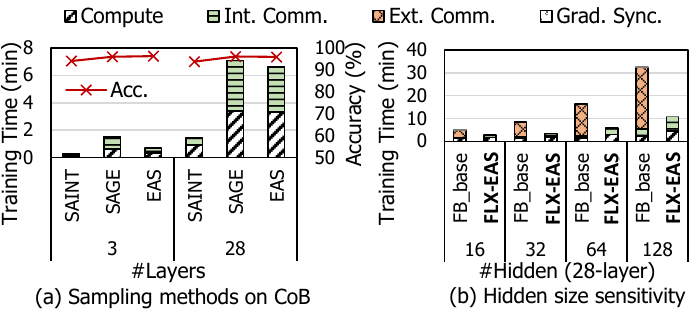}
 \vspace{-7mm}
 \caption{(a) Comparison of sampling methods implemented on \cobatch. (b) Hidden size sensitivity of \thiswork (GRD) on Reddit dataset.}
 \vspace{-3mm}
 \Description[]{Sampling and hidden size sensitivity test.}
 \label{fig:orth_hidden}
\end{figure}

\textbf{Hidden dimensions.}
\cref{fig:orth_hidden}b shows the hidden dimension size sensitivity of \thiswork on the Reddit dataset with 28-layer GNN.
Overall, \thiswork (FLX-EAS) consistently provides 1.85-2.97$\times$ speedup over FB\_base.
As the hidden dimension size grows, the portion of external communication increases, which leads to more speedup of \thiswork.

 \begin{figure}[t]
\centering
 \includegraphics[width=\columnwidth]{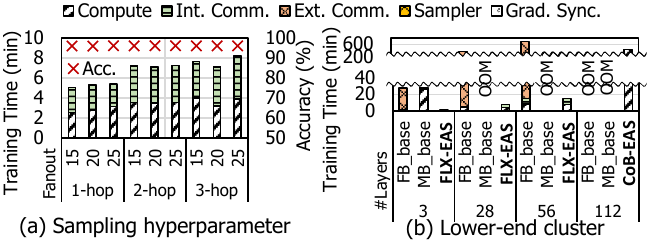}
 \vspace{-7mm}
 \caption{(a) Sensitivity of \intersampling on hyper-parameter with Reddit. (b) Training time breakdown on a lower-end cluster with Reddit.}
 \vspace{-5mm}
 \Description[]{Hyperparameter sensitivity and lower-end cluster test.}
 \label{fig:param_lowend}
\end{figure}

\textbf{Sampling hyperparameters.}
\cref{fig:param_lowend}a illustrates how the hyperparameters of further hop and larger fanout change training time and accuracy of \intersampling.
We tested the 1- to 3-hop settings with a 15- to 25-fanout setting, which are widely used fanout values in sampling methods.
All experiments are conducted in the full-batch mode with \intersampling (FLX-EAS).
As we increase the maximum hops, the training time increases up to 2-hop with a 25-fanout setting.
On the other hand, all settings show similar accuracy, which means that \intersampling can reach full accuracy even with 1-hop with a 15-fanout (default).

\textbf{Lower-end cluster.}
The ability of \thiswork being able to free transit between full-batch and mini-batch shines better with low-end hardwares, especially with limited memory.
We equipped a lower-end cluster with two servers, each with four RTX 2080Ti (11GB device memory), connected with 1GbE external bandwidth.
As illustrated in~\cref{fig:param_lowend}b, in various layer settings with a hidden size 64, \thiswork provides 29.07-44.64$\times$ speedup on Reddit compared to FB\_base and MB\_base.
MB\_base suffers from GPU OOM due to severe redundancy in 28-112 layers, even with a batch size of 1.
FB\_base also cannot handle Reddit with 112 layers, but \thiswork switches to \cobatch with \intersampling, which becomes the only available framework.

\subsection{Analysis of Memory Consumption}

 \begin{figure}[t]
\centering
 \includegraphics[width=\columnwidth]{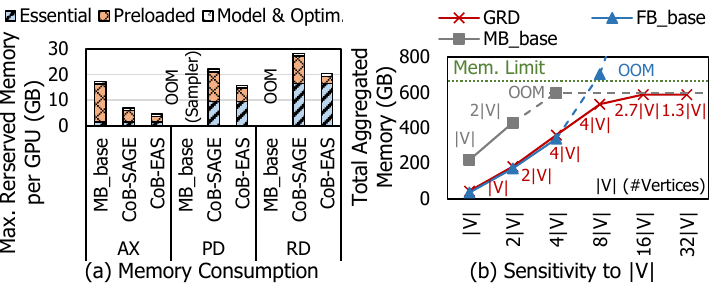}
 \vspace{-7mm}
 \caption{(a) Memory consumption analysis of MB\_base and \thiswork. (b) Memory consumption of the baselines (FB\_base, MB\_base) and \thiswork (GRD) when the graph size becomes large.}
 \vspace{-5mm}
 \Description[]{Memory-related analysis.}
 \label{fig:memory_analysis}
\end{figure}

\cref{fig:memory_analysis}a shows the memory consumption of the baseline MB\_base (with SAGE) and \cobatch (CoB-SAGE, CoB-EAS) in a deep GNN of 56 layers and 256 hidden dimensions with the same batch size.
\Cobatch significantly reduces the redundant memory usage of MB\_base, and EAS additionally suppresses it.
\cref{fig:memory_analysis}b gives further insights about the memory usage of the baselines and \thiswork.
We generated random synthetic graphs of various numbers of vertices, following the average degree of the Reddit dataset.
We used the 28-layer model, and colored $n$|V| indicates the maximum batch size.
GRD consumes slightly more memory than FB\_base from \mempreload.
When the graph becomes larger, FB\_base fails to handle it due to memory limit.
In contrast, GRD keeps training by transitioning to mini-batch training with \cobatch, which is the advantage of a unified framework.
On the other hand, MB\_base consumes significantly larger memory than GRD with the same batch size due to redundant memory usage.
Due to the neighbor explosion, MB\_base fails to train large graphs over 4|V|.
However, GRD can train them through \cobatch and \intersampling.

\subsection{Full-Batch Sampling/Staleness Methods}
\label{sec:stale_sampling}

\begin{figure}[t]
    \centering
    \includegraphics[width=\columnwidth]{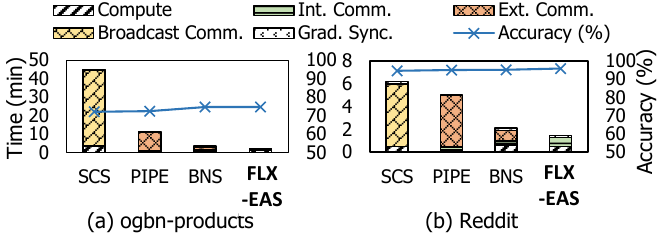}
    \vspace{-7mm}
    \caption{Training time breakdown and accuracy of \thiswork with \intersampling compared to sampling- and staleness-based full-batch training methods.}
    \vspace{-3mm}
    \Description[]{Comparison with full-batch-based sampling and staleness methods.}
    \label{fig:sampling_stale}
\end{figure}

Some prior works have introduced sampling-~\cite{bns_gcn} or staleness-~\cite{pipegcn,sancus} based methods to address the massive communication overhead of full-batch training.
Sancus~\cite{sancus} (SCS) uses the staled historical embeddings.
PipeGCN~\cite{pipegcn} (PIPE) overlaps the communication by computation through pipelining and uses the staled embeddings.
BNS-GCN~\cite{bns_gcn} (BNS) samples the communication to be around 10\% and uses stale activations.
Even though \thiswork does not involve any staleness, we directly compared \mempreload with \intersampling (FLX-EAS) to the methods described above. 
We used the Products and Reddit dataset using the open-sourced code of~\cite{sancus, pipegcn, bns_gcn} in \cref{fig:sampling_stale}.
We used 3 layers containing 2 GCN layers and a single FC layer with a hidden size of 256.
We fixed a learning rate of 0.01 and a dropout of 0.0 following~\cite{sancus}.

As shown in \cref{fig:sampling_stale}, the communication volume of broadcast-based SCS is large because it is proportional to the square of the total number of vertices.
By contrast, the communication volume of all-to-all-based methods (PIPE, BNS, and FLX-EAS) is proportional to the total number of edges.
Therefore, those works provide more speedup over SCS on the Product dataset, which has a lower graph density (\#Edges/\#Vertices) than the Reddit dataset.
\thiswork (FLX-EAS) shows sufficient speedup over all prior works, with values ranging from 1.45$\times$ to 28.04$\times$ because they suffer from the slow inter-server communication.
Further, \thiswork provides the same or more stable accuracy than the baselines.

\subsection{Accuracy/Actual Total Error Comparison}
\label{sec:acc_compare}

We compared the accuracy of non-sampled FB\_base and \intersampling (FLX-EAS) with 3-layer GraphSAGE model in~\cref{tab:acc_compare}.
\thiswork provides comparable accuracy to the non-sampled training by keeping the intra-server dependency and due to the effect of a large batch, as discussed in \cref{sec:err_bound}.

\begin{figure}[t]
\centering
 \includegraphics[width=.92\columnwidth]{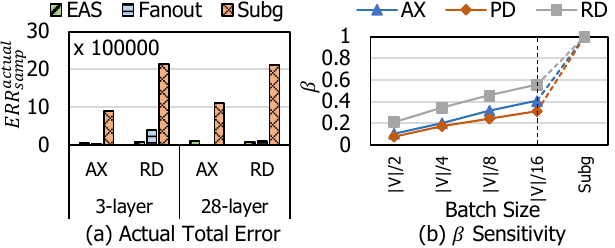}
 \vspace{-3.5mm}
 \caption{(a) Actual values of total errors from samplings on the four-server setting and (b) sensitivity of $\beta$ on batch sizes.}
 \vspace{-3mm}
 \Description[]{Error bound and actual error analysis.}
 \label{fig:err_bound_analysis}
\end{figure}

To have a deeper insight, we also directly measured the actual total error coming from \intersampling, fanout sampling, and subgraph sampling in \cref{fig:err_bound_analysis}a by measuring the total value of~\cref{eq:error_lemma_layers} in \cref{sec:err_bound} using the trained weights.
Due to large $\beta$ (i.e., the portion of boundary vertices), subgraph sampling shows substantially larger total errors than others.
\Intersampling obtains the advantage of the large batch (small $\beta$) and provides low actual errors.
Fanout sampling also shows low errors but suffers from increased errors in a high average degree dataset (i.e., Reddit).
This is because fanout sampling restricts fixed fanout for all vertices.
For Arxiv, fanout sampling shows slightly lower errors than \intersampling, but it is natural because the average degree of Arxiv is around $7$, much smaller than the used fanout 25.
The sensitivity of $\beta$ on batch sizes ranges from $0.08$ to $0.56$, as shown in \cref{fig:err_bound_analysis}b.
$\beta$ is smaller than 0.6, even using 16 servers ($|V|/16$), so \intersampling can provide stable convergence in more extensive cluster settings.

\section{Discussion}
\subsection{Quantitative Memory Analysis}
We approximately formulated \thiswork's additional memory usage as follows.
When scaling the number of servers, the total aggregated additional memory usage increases, but the additional memory usage per GPU is stable due to the effect of utilizing multiple servers/GPUs and \intersampling.

For formulation, we set some notations and assumptions as below.
We use two number of servers ($N_{s_1}$ and $N_{s_2}$) and $N_{s_2}=m \times N_{s_1}$.
We also assume that we already applied \intersampling of $h=1$ (1-hop) for brevity.
$H$ indicates the hidden size, and we set the initial feature size as the same as the hidden size.
$1<\alpha_{1}<\alpha_{2}$ are graph- and \#servers-dependent per-layer expansion factor.
$\alpha_2$ is larger than $\alpha_1$ because a graph expands more when \#target vertices is small.
$\alpha_{2} = n \times \alpha_{1}$ where $n << m$ due to the min-cut partitioning algorithm (i.e., METIS).
The dependency graph size grows $\alpha$ for 1-hop because we applied \intersampling.

The memory usage/GPU of \mempreload using \intersampling ($h=1$) with $s_1$ servers ($M^{s_1}_{\mathit{preload}}$) is formulated as below ($2 \times$ for forward/backward):
\begin{align}
M^{s_1}_{\mathit{preload}} = 2\times\frac{VHL \alpha_1}{N_{s_1}N_g}.
\end{align}
Now, we further formulate how the additional memory usage changes when increasing the number of servers to $s_2$.
When we compare the total aggregated memory usage of the whole server clusters, this can be formulated as follows:
\begin{align}
2 \times VHL(\alpha_2-\alpha_1) > 0~~(\because \alpha_1 < \alpha_2).
\end{align}
This indicates that the total aggregated memory usage increases with larger \#servers.
However, in terms of each GPU, we can compare the additional memory usage per GPU with $s_1$ and $s_2$ servers ($M^{s_2}_{preload} / M^{s_1}_{preload}$) as follows:
\begin{align}
\frac{M^{s_2}_{preload}}{M^{s_1}_{preload}} = \frac{N_{s_1}\alpha_{2}}{N_{s_2}\alpha_{1}} = \frac{n}{m} < 1.
\end{align}
When using the min-cut partitioning and \intersampling, $n << m$ usually holds.
Therefore, the additional memory usage per GPU is stable when scaling the number of servers.

\begin{table}[t]
    \centering
    \caption{Test Accuracy of FB\_base and FLX-EAS} \label{tab:acc_compare}
    \vspace{-3mm}
    \resizebox{\columnwidth}{!}
    {
    \setlength{\tabcolsep}{3pt}
    \begin{tabular}{cccccccccc}
    \toprule
         Dataset& Arxiv & Products & Reddit & IGB (S) & IGB (M) & Papers \\
         \midrule
         FB\_base & 69.03\% & 75.69\% & 96.33\% & 74.63\% & 73.49\% & OOM \\
         FLX-EAS & 69.21\% & 75.31\% & 96.33\% & 74.70\% & 73.55\% & 60.84\%   \\
        \bottomrule
    \end{tabular}
    } \vspace{-3mm}
\end{table}

\subsection{Application on High-End Systems}
While we tested \thiswork on a four-server cluster with fast Infiniband (32Gbps) and NVLink Bridges (112GB/s), recent high-end systems (e.g., DGX-A100) often introduce much faster inter-server connection (e.g., 200Gbps Infiniband) and intra-server connection (e.g., 600GB/s NVLink Switch).
We believe \thiswork would provide speedup in such systems because their intra-/interbandwidth ratio (600GB/200Gb) does not significantly differ from ours (112GB/32Gb).
\thiswork mainly benefits from trade-offing intra-server communication with inter-server communication, so the speedup will persist. 

\section{Related Work}

\textbf{Large graphs and deep GNNs.} There have been attempts to increase the layer of GNN, as in traditional CNN models~\cite{resnet, preact}.
It was shown that naively increasing GNN layers suffer from over-smoothing issues.
Therefore, \cite{deepgcn, deepergcn} applied residual connection and preactivation to deep GNNs to mitigate it.
These deep GNNs have shown superior accuracy in large graph datasets~\cite{ogb, oag}.
To meet memory limitations, many previous~\cite{graphsage,fastgcn, gnnlab, giant_graph} works have used a sampling method in GNN training.  
\cite{salient, gpu_oriented} further tried to reduce the sampling overhead by overlapping the data transfer with computation.
\cite{clustergcn, graphsaint} use the subgraph structure of graphs and find local neighbors.
Some works also focused on the overhead of preparing data for training and developed a cache policy~\cite{bgl, pagraph, gnnlab} or sampling~\cite{bns_gcn} to reduce feature retrieving costs.

\textbf{GNN frameworks.} There are many recent works~\cite{aligraph, distdgl, distdglv2, neugraph, pagraph, dorylus, sancus, cagnet, distmemoryfg} which aim to perform distributed training in a large graph.
Most frameworks focus on how to handle the dependencies.
\cite{aligraph} separately stores attributes of graphs in different workers and caches frequent neighbors.
\cite{agl} designed k-hop neighborhood for each node and therefore enables parameter-server~\cite{parameter_server} based training.
However, \cite{aligraph, agl} only support CPU-based GNN training.
\cite{distdgl, distdglv2} attempt to handle the dependency among vertices with caching and, therefore, suffer from the redundant computation.
On the other hand, ~\cite{roc, pipegcn, cagnet, sancus, distmemoryfg, dorylus} attempt to solve this dependency with communication.
~\cite{cagnet} uses sequential broadcast methods to handle the vertex dependency among distributed workers.
~\cite{sancus} further accelerates it by using historical embeddings to avoid communication.
~\cite{dorylus} considers low-end servers, but it is limited to CPU servers.
While focusing on the limitation of each approach, a hybrid approach was proposed ~\cite{neutronstar}, but with an unrealistic assumption that each server has only one GPU.
\cite{neugraph} combined graph computation optimization with data partitioning and scheduling, and \cite{distmemoryfg} adopted hypergraph partitioning model to derive optimized communication operations.
However, the prior works~\cite{neugraph, pagraph, gnnlab} have not carefully considered a multi-server and multi-GPU environment, which has a large gap between internal and inter-server bandwidth.
To our knowledge, this is the first approach to consider a multi-server cluster environment.
Multi-GPU GNN training frameworks~\cite{dsp, quiver, neutronorch, legion, hongtu} are also widely used in GNN training.
However, they mainly discuss based on the multi-GPU concept while not considering the multi-server setup.
\thiswork provides novel strategies considering the intra-/inter-server link bandwidth difference, including a sampling strategy.
The contributions of multi-GPU GNN training could be applied to \thiswork in terms of intra-server multi-GPUs.

\textbf{Caching in GNN training.}
Many mini-batch training frameworks~\cite{wholegraph, quiver, pagraph,icache} propose caching strategies for GNN training.
They mainly focus on caching the initial features of mini-batches~\cite{quiver, pagraph}, and introduce prefetch or dedicated caching algorithms~\cite{pagraph, icache}.
\cite{wholegraph} provides shared GPU memory with multiple GPUs to use GPU memory as a cache.
However, these works still face the redundancy of mini-batch because they do not consider the characteristics of multi-server clusters.
In other words, as they target an individual GPU, they suffer from redundancy issues and need to cache duplicate initial features.
On the other hand, \thiswork proposes a kind of server-level caching through fast intra-server communication, which is one of the main characteristics of recent clusters.
As a result, \thiswork significantly suppresses the redundancy and increases the scalability and throughput.   

\textbf{GNN training on limited environments.}
While \thiswork targets a multi-server cluster with GNN training, many researches~\cite{marius, hongtu, helios, ginex, ducati} suggest GNN training for limited environments.
As recent graphs have become extremely large, those works provide methods to address the sampling bottleneck of mini-batch training.
\cite{marius, helios, ginex} mainly focus on efficiently utilizing SSDs when generating mini-batch subgraphs from SSDs.
\thiswork supports both full-/mini-batch training, so these works could be orthogonally applied when training is conducted with much larger graphs.

\textbf{Graph processing and GNN inference.}
While \thiswork mainly discusses GNN training, there exists a lot of graph processing~\cite{graphicionado, extrav, piccolo_cal} and GNN inference~\cite{hygcn, awb_gcn, sgcn, snf, snf_cal} research targeting various environments.
\cite{graphicionado} proposes an efficient graph accelerator structure for graph processing and \cite{piccolo_cal} utilizes in-memory scatter-gather for addressing the random access of graph processing for acceleration.
\cite{extrav} emphasizes the importance of near-storage processing for graph processing with a dedicated accelerator.
For GNN inference, many works~\cite{hygcn, awb_gcn} propose various types of accelerators to reduce the inference latency.
\cite{snf, snf_cal} focus on efficiently utilizing on-chip memory and \cite{sgcn} exploits the sparsity of GNN activations while inferencing.

\section{Conclusion}
To the best of our knowledge, \thiswork is the first distributed GNN training framework to fully consider the characteristics of multi-server clusters.
By utilizing the chance provided by multi-server clusters, \thiswork successfully addresses the inter-server communication bottleneck of distributed training.
Additionally, \thiswork mitigates the redundancy issue of mini-batch training by aggregating multiple GPUs in a server with fast intra-server links.
Lastly, \thiswork proposes a novel sampling method, suited for multi-server clusters, which significantly suppresses the neighbor explosion.
As a result, \thiswork is a fast GNN training framework that provides state-of-the-art training throughput.

\begin{acks}
\begin{sloppypar}
This work was supported by Samsung Advanced Institute of Technology, Samsung Electronics Co., Ltd. (IO230223-05124-02), and the National Research Foundation of Korea (NRF) grant funded by the Korean government (MSIT) (2022R1C1C1011307, 2022R1C1C1008131).
This work was partly supported by an IITP grant funded by the Korean Government (MSIT) (No. RS-2020-II201361).
Seongyeon Park assisted in revising the paper, and we thank her for the contribution.    
\end{sloppypar}
\end{acks}

\section*{Data Availability Statement}
The artifact of \thiswork is available at~\cite{granndis_artifact}.
The most recent version is kept updated at \url{https://github.com/AIS-SNU/GraNNDis_Artifact}.
The artifact contains some \thiswork code and evaluation runners for distributed environments.
Also, it includes an environment setup script.
For the details of the artifact and the expected results, please refer to \cref{sec:artifact}.

\appendix
\section{Artifact Appendix}
\label{sec:artifact}
\subsection{Abstract}
We provide \thiswork's source code and additional code for setup and execution.
For the most recent version of \thiswork's description, please refer to the up-to-date artifact link in GitHub~\cite{granndis_artifact}.

\subsection{Artifact Summary}

{\small
\begin{itemize}
  \item {\bf Algorithm:} Distributed graph neural network training.
  \item {\bf Program: } Python with PyTorch and DGL.
  \item {\bf Dataset: } Sample graph datasets (ogbn-arxiv/products~\cite{ogb}, Reddit~\cite{reddit})
  \item {\bf Run-time environment: } Ubuntu 22.04 or higher.
  \item {\bf Hardware: } Multiple servers and each server has multiple GPUs. The internal server bandwidth (e.g., NVLink) is recommended to be much higher than the external server bandwidth (e.g., Ethernet).
  \item {\bf Metrics: } Execution time and accuracy.
  \item {\bf Recommended disk/memory space: } 1TB/256GB.
  \item {\bf Preparation time: } For software setting, it takes less than an hour.
  \item {\bf Experiment time: } It takes less than two hours for sample datasets.
  \item {\bf Code licenses: } MIT license.
  \item {\bf Archived DOI: } \url{https://doi.org/10.5281/zenodo.12738844}
  \item {\bf Up-to-date artifact: } \url{https://github.com/AIS-SNU/GraNNDis_Artifact}
\end{itemize}
}

\subsection{Description}
\subsubsection{How to access}
Please access the artifact through the archived DOI~\cite{granndis_artifact} or the up-to-date artifact link.

\subsubsection{Hardware dependencies}
Muti-server environment, and each server is equipped with multiple GPUs. Internal server interconnect (e.g., NVLink) is much faster than external server interconnect (e.g., 10G Ethernet).

\subsubsection{Software dependencies}
CUDA/CuDNN 11.8 Setting (Make sure to include CUDA paths). Anaconda Setting. The NFS environment has more than two servers, and each server has multiple GPUs.
Servers must be accessible by SSH connection without password using \texttt{ssh-copy-id} (e.g., ssh [user]@[server]).

\subsubsection{Datasets}
For the artifact evaluation, we use three sample datasets (Arxiv, Reddit, and Products), which are widely used and easily accessible.
In the following subsection, the datasets will be automatically downloaded.

\subsection{Installation}

\subsubsection{Software installation}
Before installation, pre-install the Anaconda environment manager. Then execute  the following:
\begin{verbatim}
  $ conda update -y conda
  $ conda create -n granndis_ae python=3.10 -y
  $ conda activate granndis_ae
  $ conda install -c conda-forge -c pytorch -c nvidia \
    -c dglteam/label/th21_cu118 \
    --file conda-requirements.txt -y
  $ pip install -r pip-requirements.txt
\end{verbatim}

\subsubsection{Dataset preparation}
\begin{verbatim}
  $ cd Codes
  $ chmod +x brief_masking_test.sh
  $ ./brief_masking_test.sh
\end{verbatim}
While running the script, you may be required to type \texttt{y} to download a dataset. The logs will be saved in \texttt{Codes/masking\_test\_logs/} if the tests are successfully conducted.

\subsection{Experiment workflow}
Some users are unfamiliar with the distributed training procedure, so we provide simple distributed experiment launchers at \texttt{AE/*.py}.
Before reproducing, users must change the configuration fields in the config file (\texttt{AE/configs.py}).

\begin{verbatim}
global_configs = {
    'env_loc': '(...)/envs/granndis_ae/bin/python',
    'runner_loc': '(...)/Codes/main.py',
    'our_runner_loc': '(...)/Codes/our_main.py',
    'workspace_loc': '(...)/GraNNDis_Artifact/',
    'data_loc': '~/datasets/granndis_ae/',
    'num_runners': 2,
    'gpus_per_server': 4,
    'hosts': ['192.168.0.5', '192.168.0.6']
}
\end{verbatim}

After modification, the following commands will show the artifact evaluation results.

\begin{verbatim}
  $ sh run_ae.sh # run AE scripts
  $ sh parse_ae.sh # parse AE results
\end{verbatim}

\subsection{Evaluation and expected results}
The results will be saved in the \texttt{AE*\_results.log}.
All FLX, CoB (with SAGE sampling), and EAS would generally show significant speedup over the baseline optimized full-batch training because GraNNDis minimizes the slow external server communication (\cref{ae:1:throughput}). EAS (FLX-EAS) is expected to show more speedup than FLX, especially in larger datasets, such as Products. EAS usually shows higher speedup than CoB (especially in larger datasets) while providing comparable accuracy, as shown in \cref{ae:2:accuracy} (accuracy result).
Please note that the result can fluctuate when the inter-server connection is shared with the cluster's NFS file system. In this case, running multiple trials will show the trend mentioned above.
The following are the example results of running the above procedure on the authors' remote machine.

\subsubsection{Throughput (execution time)}
\label{ae:1:throughput}
{\small
\begin{verbatim}
+-------------------------------------------------------+
|              Throughput Results for Arxiv             |
+--------+------------------+-----------------+---------+
| Method | Total Time (sec) | Comm Time (sec) | Speedup |
+--------+------------------+-----------------+---------+
| Opt_FB |      15.40       |       9.85      |   1.00  |
|  FLX   |       8.60       |       2.37      |   1.79  |
|  CoB   |       8.78       |       2.58      |   1.75  |
|  EAS   |      11.67       |       3.70      |   1.32  |
+--------+------------------+-----------------+---------+
+-------------------------------------------------------+
|             Throughput Results for Reddit             |
+--------+------------------+-----------------+---------+
| Method | Total Time (sec) | Comm Time (sec) | Speedup |
+--------+------------------+-----------------+---------+
| Opt_FB |      449.27      |      422.35     |   1.00  |
|  FLX   |      87.55       |      49.67      |   5.13  |
|  CoB   |      90.44       |      49.98      |   4.97  |
|  EAS   |      75.16       |      40.34      |   5.98  |
+--------+------------------+-----------------+---------+
+-------------------------------------------------------+
|            Throughput Results for Products            |
+--------+------------------+-----------------+---------+
| Method | Total Time (sec) | Comm Time (sec) | Speedup |
+--------+------------------+-----------------+---------+
| Opt_FB |      79.67       |      69.15      |   1.00  |
|  FLX   |      20.03       |       6.36      |   3.98  |
|  CoB   |      21.85       |       8.33      |   3.65  |
|  EAS   |      18.23       |       5.78      |   4.37  |
+--------+------------------+-----------------+---------+
\end{verbatim}
}

The results show that the optimized full-batch training baseline (Opt\_FB) suffers from communication overhead, while \mempreload (FLX)/\cobatch (CoB) addresses such an issue through server-wise preloading. \Intersampling (EAS) further accelerates the training through server boundary-aware sampling. This trend becomes vivid in larger datasets (Reddit and Products).

\subsubsection{Accuracy}
\label{ae:2:accuracy}
{\small
\begin{verbatim}
+--------------------------------------+
| Accuracy Comparison (FB vs. FLX-EAS) |
+--------+-------+--------+------------+
| Method | Arxiv | Reddit |  Products  |
+--------+-------+--------+------------+
|   FB   |  0.69 |  0.96  |    0.76    |
|  EAS   |  0.69 |  0.96  |    0.76    |
+--------+-------+--------+------------+
\end{verbatim}
}

As \intersampling (EAS) only targets sample server boundary vertices, contributing to acceleration, it achieves comparable accuracy to the original full-batch training.

\subsection{Notes}
For detailed arguments and distributed launch processes, please refer to the additional guidelines of the readme in the artifact.

\clearpage

\bibliographystyle{ACM-Reference-Format}
\bibliography{pact24_granndis}

\end{document}